\documentclass[10pt,twocolumn,letterpaper]{article}

\usepackage{cvpr}              

\usepackage{multirow}
\usepackage{color}
\usepackage{fontawesome}
\usepackage{colortbl}
\usepackage{dsfont}
\usepackage{stackengine}
\usepackage{times}
\usepackage{graphicx}
\usepackage{amsmath}
\usepackage{amssymb}
\usepackage{multirow}
\usepackage{mathtools,xparse}
\usepackage{subcaption}
\usepackage{stmaryrd}
\usepackage{placeins}
\usepackage{pifont}
\usepackage{tabularx}
\usepackage{mdframed}
\usepackage{nopageno}
\usepackage{makecell}
\usepackage{arydshln}
\usepackage{booktabs}
\usepackage{enumitem}
\usepackage[skip=1ex,font=small,labelsep=period]{caption}
\usepackage{tabu}
\usepackage{etoolbox}
\usepackage{verbatim}
\usepackage{siunitx}
\usepackage{xspace}
\usepackage[table,dvipsnames]{xcolor}
\usepackage{epsfig}
\usepackage[noend]{algpseudocode}
\usepackage{algorithm}
\usepackage{lipsum}
\usepackage[accsupp]{axessibility} 

\newcommand\blfootnote[1]{%
  \begingroup
  \renewcommand\thefootnote{}\footnote{#1}%
  \addtocounter{footnote}{-1}%
  \endgroup
}

\usepackage[pagebackref,breaklinks,colorlinks]{hyperref}

\definecolor{wincolor}{rgb}{0.95, 0.2, 0.2}

\robustify\bfseries 
\newcolumntype{Y}{>{\centering\arraybackslash}X}

\makeatletter
\let\oldabs\abs
\def\abs{\@ifstar{\oldabs}{\oldabs*}}
\let\oldnorm\norm
\def\norm{\@ifstar{\oldnorm}{\oldnorm*}}
\makeatother

\definecolor{mygray}{gray}{.95}
\definecolor{headergray}{gray}{.85}
\definecolor{Gray}{gray}{0.85}
\newcolumntype{g}{>{\columncolor{Gray}}c}

\makeatletter
\def\BState{\State\hskip-\ALG@thistlm}
\makeatother

\algnewcommand\algorithmicforeach{\textbf{for each}}
\algdef{S}[FOR]{ForEach}[1]{\algorithmicforeach\ #1\ \algorithmicdo}

\newcommand{\customparagraph}[1]{\noindent \textbf{#1}}
\newcommand{\customsubparagraph}[1]{\noindent \textit{#1}}

\newcommand*\matr{\mathbf}
\newcommand*\trans{\text{T}}
\newcommand*\Instance{h}

\newcommand*\Instances{\mathcal{I}}
\newcommand*\Points{\mathcal{P}}

\newcommand*\Point{\mathbf{p}}

\newcommand*\DistanceFunction{\phi}

\newcommand*\Method{proposed method }
\algrenewcommand\algorithmicrequire{\textbf{Input:\phantom{ll}}}
\algrenewcommand\algorithmicensure{\textbf{Output:}}

\let\emptyset\varnothing

\usepackage[capitalize]{cleveref}
\crefname{section}{Sec.}{Secs.}
\Crefname{section}{Section}{Sections}
\Crefname{table}{Table}{Tables}
\crefname{table}{Tab.}{Tabs.}

\def\cvprPaperID{6511} 
\def\confName{CVPR}
\def\confYear{2023}

\begin{document}

\title{Finding Geometric Models by Clustering in the Consensus Space
}

\author{Daniel Barath$^1$, Denys Rozumnyi$^{2,1}$, Ivan Eichhardt$^{3, 4}$, Levente Hajder$^4$, Jiri Matas$^2$\\
$^1$Computer Vision and Geometry Group, ETH Zurich, Switzerland, \\
$^2$VRG, Faculty of Electrical Engineering, CTU in Prague, Czech Republic, \\
$^3$TMEIC Corporation Americas, Roanoke, VA, USA \,
$^4$E\"otv\"os Lor\'and University, Budapest, Hungary
}

\maketitle

\begin{abstract}
We propose a new algorithm for finding an unknown number of geometric models, \eg, homographies. 
The problem is formalized as finding dominant model instances progressively without forming crisp point-to-model assignments. 
Dominant instances are found via a RANSAC-like sampling and a consolidation process driven by a model quality function considering previously proposed instances. 
New ones are found by clustering in the consensus space.
This new formulation leads to a simple iterative algorithm with state-of-the-art accuracy while running in real-time on a number of vision problems -- at least two orders of magnitude faster than the competitors on two-view motion estimation.  
Also, we propose a deterministic sampler reflecting the fact that real-world data tend to form spatially coherent structures. 
The sampler returns connected components in a progressively densified neighborhood-graph. 
We present a number of applications where the use of multiple geometric models improves accuracy. 
These include pose estimation from multiple generalized homographies; trajectory estimation of fast-moving objects; and we also propose a way of using multiple homographies in global SfM algorithms.
Source code: \url{https://github.com/danini/clustering-in-consensus-space}.
\end{abstract}

\section{Introduction}

Robust multi-instance model fitting is the problem of interpreting a set of data points as a mixture of noisy observations stemming from multiple instances of geometric models. 
Examples for such a problem are the estimation of plane-to-plane correspondences (\ie, homography matrices) in two images, and the retrieval of rigid motions in a dynamic scene captured by a moving camera.
In the state-of-the-art algorithms, finding an unknown number of model instances is achieved by clustering the data points into disjoint sets, each representing a particular model instance. 
Robustness is achieved by considering an outlier model. 

Multi-instance model fitting has been studied since the early sixties. 
The Hough-transform~\cite{vc1962method,illingworth1988survey} is perhaps the first popular method for finding multiple instances of a single class~\cite{guil1997lower,matas2000robust,rosin1993ellipse,xu1990new}. 
The RANSAC~\cite{fischler1981random} algorithm was as well extended to deal with finding multiple instances. 
Sequential RANSAC~\cite{vincent2001detecting,kanazawa2004detection} detects instances in a sequential manner by repeatedly running RANSAC to recover a single instance and, then, removing its inliers from the point set. 
The greedy approach that makes RANSAC a powerful tool for recovering a single instance becomes its drawback when estimating multiple ones.
Points are assigned not to the best but to the first instance, typically the one with the largest support, for which they cannot be deemed outliers.
MultiRANSAC~\cite{zuliani2005multiransac} forms compound hypotheses about $n$ instances.
In each iteration, MultiRANSAC draws samples of size $n$ times $m$, where $m$ is the number of points required for estimating a model instance, \eg, $m = 4$ for homographies.
Besides requiring the number $n$ of the instances to be known a priori, 
the increased sample size affects the problem complexity and, thus, the processing time severely.


\begin{figure}[t]
    \centering
    \frame{\includegraphics[trim=0 90 20 70,clip,width=0.47\columnwidth]{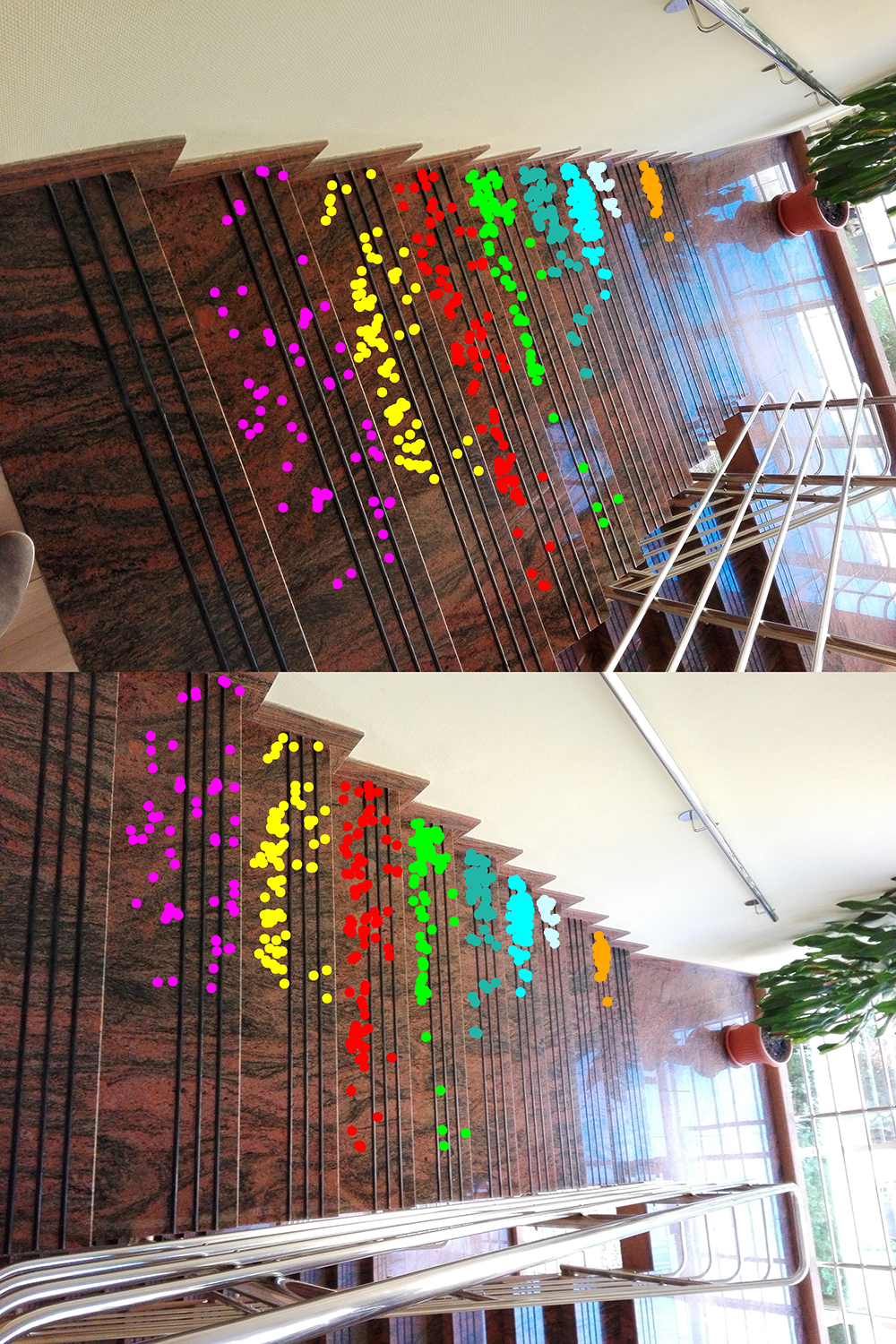}}
    \frame{\includegraphics[trim=0 90 20 70,clip,width=0.47\columnwidth]{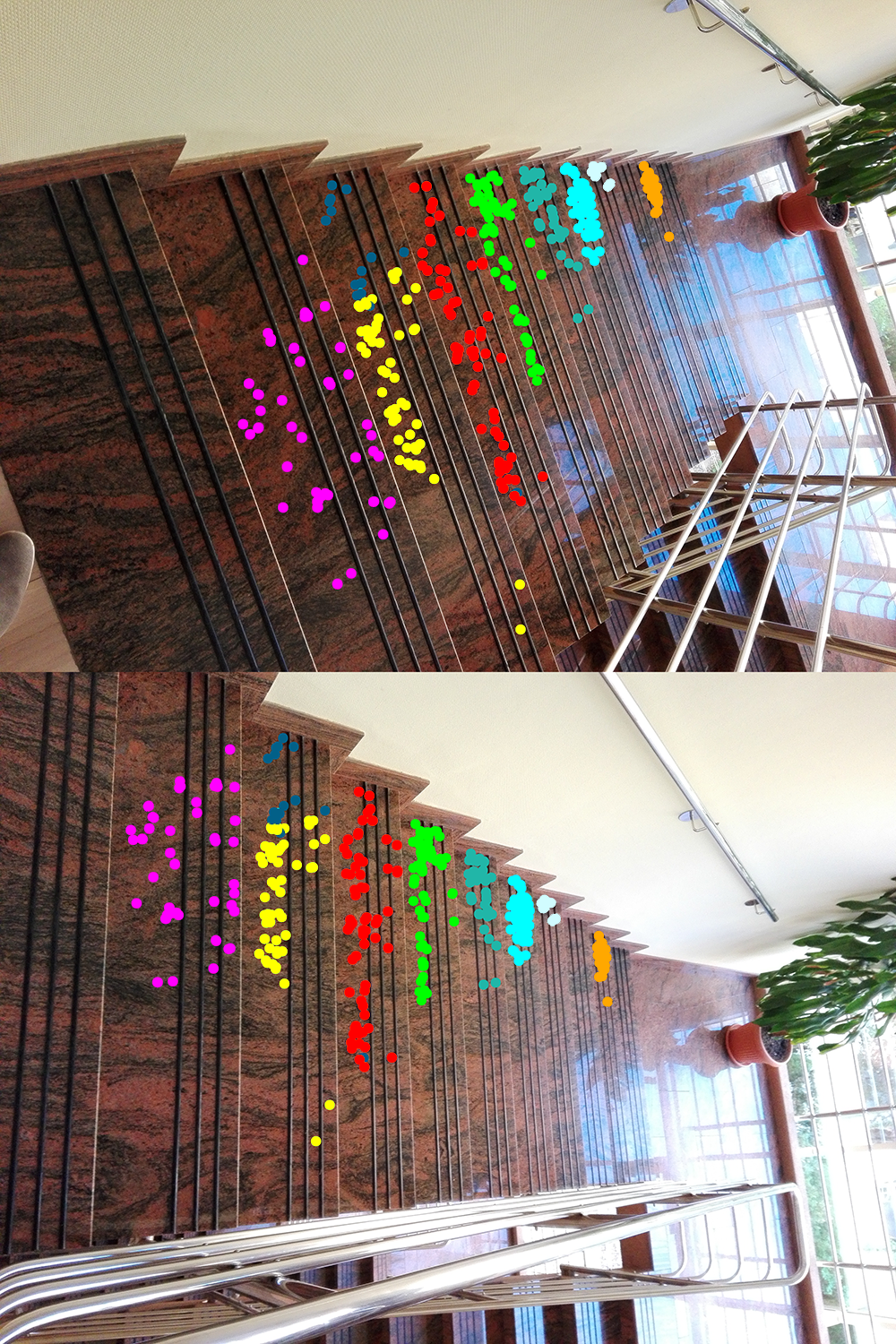}}
	\caption{ Multi-homography fitting with the \Method in $0.04$ secs (left), and with Prog-X~\cite{barath2019progressive} in $1.48$ secs (right). Prog-X is one of the fastest SOTA algorithm. Outliers are not drawn. }
	\label{fig:teaser}
\end{figure}

Modern approaches for multi-model fitting~\cite{wongiccv2011,isack2012energy,pham2014interacting,magri2014t,magri2015robust,wang2015mode,magri2016multiple,barath2018multix,amayo2018geometric} follow a two-step procedure. First, they generate many instances by repeatedly selecting minimal point sets and fitting model instances. 
Second, a subset of the hypotheses is selected interpreting the input data points the most.
This selection is done in various ways. 
For instance, a popular group of methods~\cite{isack2012energy,pham2014interacting,barath2018multix,amayo2018geometric} optimizes point-to-model assignments by energy minimization using graph labeling techniques~\cite{boykov2004experimental}. 
The energy originates from point-to-model residuals, label costs~\cite{delong2012minimizing}, and geometric priors~\cite{pham2014interacting} such as the spatial coherence of the data points.
Another group of methods uses preference analysis based on the distribution of the residuals of data points~\cite{zhang2007nonparametric,magri2014t,magri2015robust,magri2016multiple}.
Also, there are techniques~\cite{wang2015mode,wang2018searching,zhao2020quantized} approaching the problem as hyper-graph partitioning where the instances are represented by vertices, and the points by hyper-edges.

\begin{figure}[t]
    \centering
    \includegraphics[width=0.88\columnwidth]{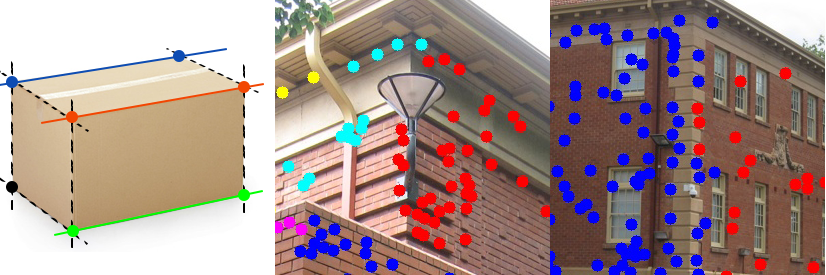}
	\caption{ \textbf{Left}: A case when assigning points to a single line (by color) prevents finding all $9$ visible instances. 
	Dashed black lines are not recovered.
	When fitting planes to $4$ out of the $7$ points, only a single plane can be found. 
	\textbf{Middle, Right}: Examples where the point-to-model assignment fails at the intersection of planes. }
	\label{fig:example_intersection}
\end{figure}

Prog-X~\cite{barath2019progressive} and CONSAC~\cite{kluger2020consac} discussed that the first, instance generation, step of the mentioned methods leads to a number of issues, \eg, the instances are generated blindly, having no information about the data at hand. 
This approach severely restricts the out-of-the-box applicability of such techniques since the user either has to consider the worst-case scenario and, thus, generate an unnecessarily high number of instances; or requires some rule of thumb, \eg, to generate twice the point number hypotheses that provides no guarantees of finding the sought instances.
Prog-X approaches the problem via interleaving the model proposal and optimization steps.
CONSAC further improves it by using a deep-learning-based guided sampling approach.

A common point of \textit{all} state-of-the-art algorithms is formalizing the multi-model fitting problem as finding disjoint sets of data points each representing a model instance.
There are two main practical issues with this assumption. 
First, in some cases, a point belongs to multiple instances and this assumption renders the problem \textit{unsolvable}, see the left image of Fig.~\ref{fig:example_intersection}.
Also, the point-to-model assignment is often unclear even if it is done by a human, especially, for points around the intersection of instances, see the right two plots of Fig.~\ref{fig:example_intersection} for examples.
%
The second issue stems from the recovery of disjoint point sets that usually requires a rather complex procedure, \eg labeling via energy minimization, that affects the run-time severely.

The \textit{main contribution} of this paper is a fundamentally new problem formulation that does not require forming crisp point-to-model assignments, \ie, a point can be assigned to multiple instances. 
This is different from the formulations used in the state-of-the-art algorithms for general multi-model fitting~\cite{isack2012energy,pham2014interacting,magri2016multiple,barath2018multix,amayo2018geometric,wang2018searching,barath2019progressive,kluger2020consac}.
This property allows the \Method to be a simple iterative algorithm and, yet, to obtain results superior to the state-of-the-art both in terms of accuracy and run-time, being \textit{real-time} on a number of problems, see Fig.~\ref{fig:teaser}, including ones where multi-model fitting algorithms generally are \textit{not} real-time, \eg, two-view motion detection.
Also, this assumption relaxes the greedy nature of sequential algorithms as the ordering in which the instances are proposed becomes unimportant.  
%
As the \textit{second} contribution, we discuss ways of exploiting multiple instances in popular applications -- Structure-from-Motion, pose estimation for generalized and pin-hole cameras, and trajectory estimation of fast-moving objects.
By considering multiple models, the accuracy is increased in almost all cases on several publicly available real-world datasets.
As the \textit{third} contribution, we propose a new sampler designed specifically for multi-instance model fitting. 
The sampler considers that real-world data tend to form spatially coherent structures.
It returns the connected components in a gradually densified neighborhood-graph. 
While several samplers exist that exploit spatial properties of the data, \eg \cite{nasuto2002napsac,barath2020magsac++}, the proposed one is \textit{deterministic}. 


\section{Iterative Clustering in the Consensus Space}

We propose a new algorithm for robust multi-instance model fitting that combines the advantages of state-of-the-art algorithms 
and, also, follows a new formulation that does not require crisp point-to-model assignments for finding the dominant model instances.

\subsection{Idea and Schematic Algorithm}

The \Method is motivated by two observations about the nature of multi-model fitting problems.
First, even though all of the state-of-the-art algorithms \cite{isack2012energy,pham2014interacting,magri2016multiple,barath2018multix,amayo2018geometric,wang2018searching,barath2019progressive,kluger2020consac} formalize the problem as a clustering where a set of data points (cluster) represents a model instance, this assumption is incorrect in a number of real-world scenes. 
Moreover, one of the primary reasons of multi-model fitting algorithms often being fairly slow stems from the optimization techniques, \eg $\alpha$-expansion in PEARL~\cite{isack2012energy}, needed to solve the point-to-model assignment problem. 

Our second observation is that multi-model fitting can usually be rephrased as the problem of finding multiple \textit{dominant} instances that are reasonably \textit{different}. 
Ideally, a \textit{dominant} instance is one that represents a real structure. 
Since this is not an algorithmically measurable property, we define being dominant as having a reasonably large support not shared with other dominant instances.
We consider instances \textit{different} if they are ``far'' on the model manifold as proposed in Multi-X~\cite{barath2018multix}.
This simple formulation allows us to avoid applying complex procedures finessing to interpret point-point, model-model, and point-model interactions.
Also, it further relaxes the greedy nature of the progressive model proposal strategy introduced in Prog-X~\cite{barath2019progressive} that enables to discover the data gradually. 
The pseudo-code of formalizing the multi-model problem as finding different dominant model instances is as follows:\\[-4mm]
\begin{algorithmic}
\Require $\Points$ -- data points
\Ensure $\Instances$ -- model instances
\State $\Instances \leftarrow \emptyset$
\While{$\neg$Terminate()}
\State $\Instances \leftarrow \Instances \; \cup \; $ FindDominantInstances($\Points$)
\While{$\neg$Convergence()}
\State $\Instances \leftarrow $ SelectUniqueInstances($\Instances$)
\State $\Instances \leftarrow $ ImproveParameters($\Instances$, $\Points$)
\EndWhile
\EndWhile
\end{algorithmic}
%

\begin{figure*}
    \centering
    \begin{subfigure}[t]{0.230\textwidth}
   	 	\centering
        \includegraphics[width=1.0\columnwidth]{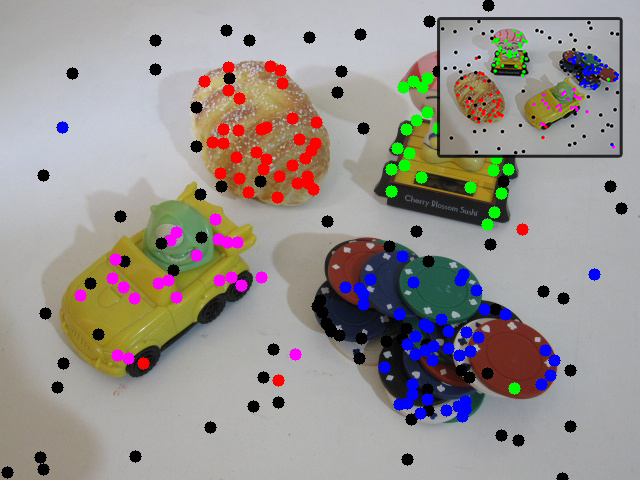}\\
        \caption{ME$_{\downarrow}$: $5.9$\%, ME$_{\uparrow}$: $5.9$\%}
    \end{subfigure}
    \begin{subfigure}[t]{0.230\textwidth}
   	 	\centering
        \includegraphics[width=1.0\columnwidth]{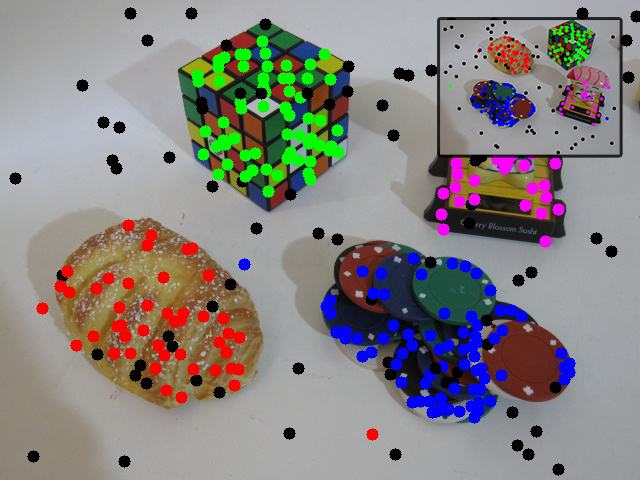}\\
        \caption{ME$_{\downarrow}$: $3.4$\%, ME$_{\uparrow}$: $3.4$\%}
    \end{subfigure}
    \begin{subfigure}[t]{0.230\textwidth}
   	 	\centering
        \includegraphics[width=1.0\columnwidth]{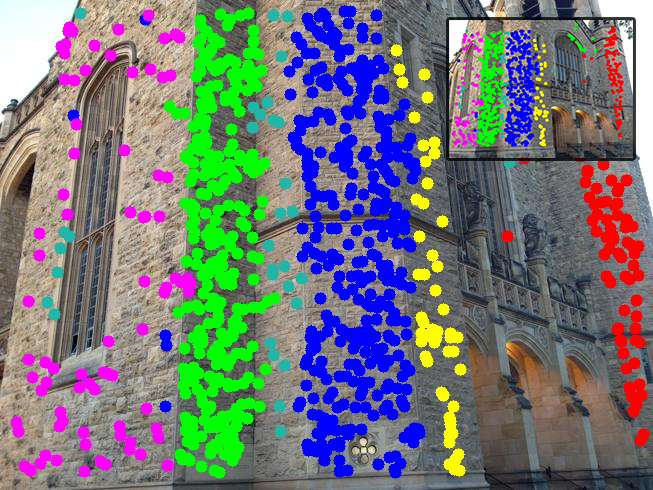}
        \caption{ME$_{\downarrow}$: $6.1$\%, ME$_{\uparrow}$: $9.4$\%}
    \end{subfigure}
    \begin{subfigure}[t]{0.230\textwidth}
   	 	\centering
        \includegraphics[width=1.0\columnwidth]{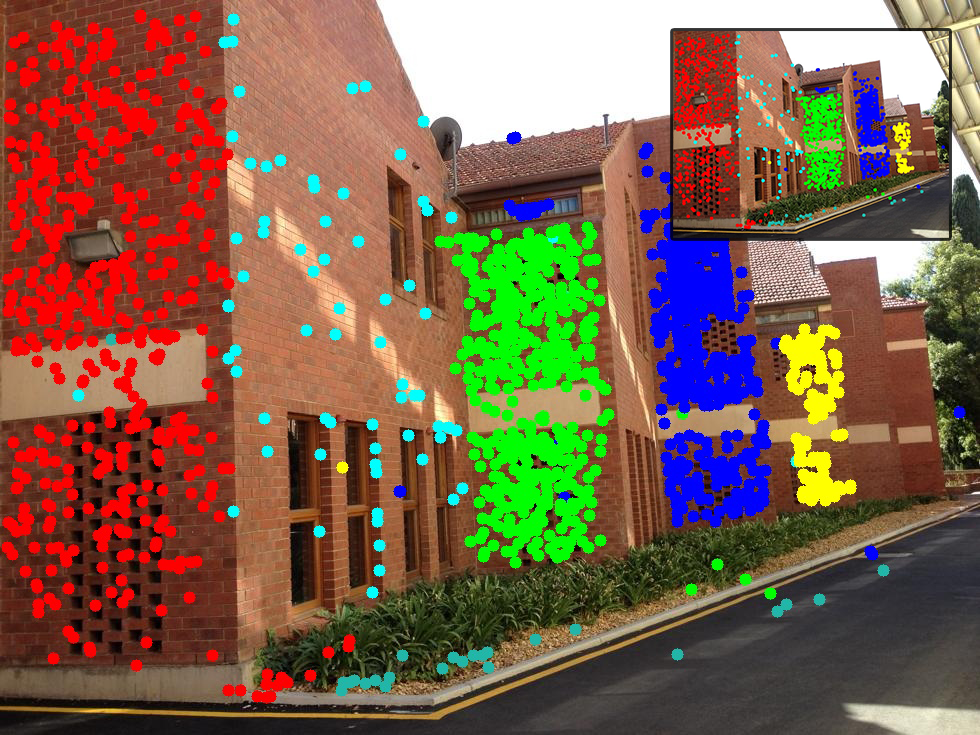}
        \caption{ME$_{\downarrow}$: $2.2$\%, ME$_{\uparrow}$: $4.1$\%}
    \end{subfigure}
	\caption{ Image pairs used for multiple two-view motion and homography estimation, and point-to-model assignments (by color) determined by assigning each point to one of the instances returned by the proposed algorithm with the minimum point-to-model residual. Black points are outliers. For each image, the highest (ME$_{\uparrow}$) and lowest (ME$_{\downarrow}$) misclassification errors in five runs are reported.
	The least accurate results are shown.
	In (a--b), the worst and best results are identical.
	In (c--d), the difference is negligible.
	The proposed method finds all sought instances, the error originates from points assigned to the wrong instance. 
	The selected scenes are the ones with the most ground truth instances to be found in the AdelaideRMF~\cite{wongiccv2011} dataset. 
	}
	\label{fig:example_results}
\end{figure*}

\subsection{Finding Dominant Model Instances}

Given a set of data points $\Points$ and a set of dominant model instances $\Instances$ proposed in earlier iterations, the objective is to find a new dominant model instance $h \in \mathbb{R}^d$ which should be included in $\Instances$, where $d \in \mathbb{R}$ is the model dimension. 
In the first iteration, $\Instances = \emptyset$.

To do so, we start similarly to RANSAC by first drawing a random sample $\mathcal{S}$ of data points.
This is done by a state-of-the-art sampler, \eg, PROSAC~\cite{chum2005matching} or P-NAPSAC~\cite{barath2020magsac++}.
Model instance $h$ is estimated from sample $\mathcal{S}$.
In order to decide about $h$ being dominant or not, we define model quality function $Q: \mathbb{R}^d \times \Points^* \times \mathbb{R} \to \mathbb{R}$ similarly as~\cite{barath2019progressive} to be calculated from the inliers of $h$ not shared with other instances in $\Instances$, where $\Points^*$ is the power set of $\Points$. Considering the RANSAC-like inlier counting, the implied quality is 
\begin{equation}
    Q_{\text{RSC}}(\Instance, \Points, \epsilon) = \sum_{\Point \in \Points} [\DistanceFunction(\Instance, \Point) < \epsilon \wedge \DistanceFunction(\Instances, \Point) \geq \epsilon],
\end{equation}
where $\epsilon \in \mathbb{R}$ is the inlier-outlier threshold and $\DistanceFunction(\Instances, \Point) = \min_{h \in \Instances} \DistanceFunction(h, \Point)$ is the minimal point-to-model residual of point $\Point$ given the kept set of dominant instances $\Instances$.
In order to use the recent advances of RANSAC, \eg the loss function of MAGSAC++~\cite{barath2020magsac++} the currently most accurate method according to a recent survey~\cite{ma2021image}, $Q_{\text{RSC}}$ is reformulated considering a continuous loss function $f$.
For practical reasons, we consider losses returning a value in-between $0$ and $1$. 
The implied quality function is
\begin{equation}
    Q_{f}(\Instance, \Points, \epsilon) = 
        |\Points| - \sum_{\Point \in \Points} \max\left( f(\Instance, \Point), 1 - f(\Instances, \Point) \right),
\end{equation}
where $f(\Instances, \Point) = \min_{h \in \Instances} f(h, \Point)$ is the minimum loss of point $\Point$ given the set of kept instances $\Instances$.
It can be easily seen that this quality function returns high score to those instances which do not share inliers with any of the instances from $\Instances$.
Otherwise, the quality is reduced according to the number and residuals of the inliers shared. 

To determine whether instance $h$ is dominant, we introduce parameter $q_{\text{min}}$, and all model instances are considered dominant where $Q_f(\Instance, \Points, \epsilon) \geq q_{\text{min}}$. 
This constraint can be interpreted as a lower bound for the number of perfectly fitting data points which are not shared with any of the instances from the maintained set in $\Instances$.\footnote{Such $q_{\text{min}}$ parameter is often used, \eg, in Structure-from-Motion algorithms (COLMAP uses $q_{\text{min}} = 15$~\cite{schonberger2016structure}). }


\subsection{Clustering in the Consensus Space}

The next step of the algorithm, after a set $\Instances$ of dominant model instances have been found, is to select a subset of $\Instances$ consisting of instances that represent different model instances and not noisy observations of the same one.
We define a model-to-model residual function $\psi: \mathbb{R}^d \times \mathbb{R}^d \to \mathbb{R}$ measuring the distance of two model instances.  

\begin{figure}[t]
    \centering
    \includegraphics[width=0.85\columnwidth]{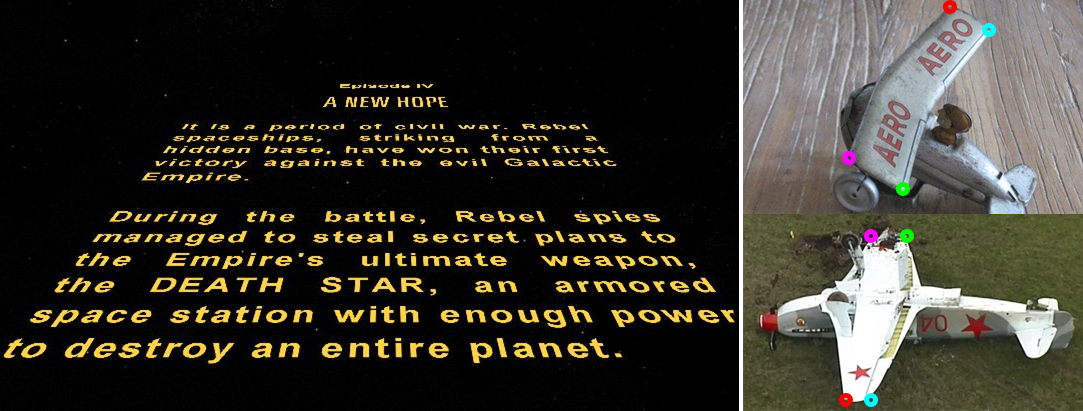}
	\caption{ Examples where converting homographies to points and back is not robust. \textbf{Left}: top two corners are mapped to the same location. Thus, three matches remain for the homography recalculation. 
	\textbf{Right}: 
	the plane flips, thus the ordering of the points 
	changes 
	and the recovered homography will be incorrect.}
	\label{fig:conversion_problem}
\end{figure}

\customparagraph{Model-to-model residual.}
Defining a 
model-to-model residual function is a challenging problem. 
In the Multi-X algorithm~\cite{barath2018multix}, it was proposed to convert the model instances to point sets.
The distance of two instances is the Hausdorff distance~\cite{rockafellar2009variational} of the point sets representing them.
This solution is however challenging, since the conversion of geometric models to point sets in a robust manner is unclear in most of the cases. 
Even for homographies, there is a number of cases when this approach simply does not work, see Fig.~\ref{fig:conversion_problem} for examples. 
Instead, we follow the strategy proposed in the T-Linkage algorithm~\cite{magri2014t} to measure the model-to-model residual as the Tanimoto distance of the preference vectors~\cite{tanimoto1958elementary} as follows:
\begin{equation}
    f_\text{T}(v_a, v_b) = \frac{\langle v_a, v_b \rangle}{||v_a||^2 + ||v_b||^2 - \langle v_a, v_b \rangle}.
\end{equation}

The preference vector of a model instance $\Instance$ is $v \in [0, 1]^n$, where $n$ is the number of input data points. Its $i$th coordinate is calculated as $v_i = 1 - f(\Instance, \Point_i)$, where $f$ is the same loss function as what is used in the previous section and $\Point_i$ is the $i$th point from $\Instances$.
Briefly, $v_i$ is zero if the point-to-model residual is greater than the inlier-outlier threshold. 
Otherwise, it is from interval $(0, 1]$. 
In this case, the Tanimoto distance measures the overlap of the inlier sets of two model instances where the inlier assignment is done in a smooth manner. 
We will call the domain of preference vectors \textit{consensus space} in the rest of the paper.

Note that, while the Tanimoto distance is not a proper metric over general vector spaces, it becomes one when the preference vector is $\in [0, +\infty)^n$~\cite{lipkus1999proof}.
This property holds in our case. 
Also note that for \textbf{0} vectors, the distance is undefined. 
In our case, this never happens since each model is fit to a minimal
sample of $m$ (= degrees of freedom)
data points which consequently have $0$ residuals. Thus, at least $m$ elements of each preference vector are $1$. 

\customparagraph{Clustering.} 
We formulate the problem of selecting different model instances as finding similar ones in $\Instances$ which are then replaced by a single instance. 
A straightforward strategy for finding similar instances is to find clusters in the consensus space defined over the preference vectors.

In general, this clustering takes place in a large dimensional space, with as many dimensions as the number of input data points.
In this particular setup however, we never have more than a few tens of instances to be clustered thanks to the iterative proposal strategy adapted from~\cite{barath2019progressive}. 
This means that the clustering is done on a few high-dimensional vectors that is very efficient with most of the clustering algorithms. 
Even if there are millions of points in the scene, a single model instance rarely has an extreme number of inliers and, thus, the indices of the non-zero elements in $v$ can be stored, making the distance calculation efficient.
In extreme cases, the min hash algorithm~\cite{broder1997resemblance} can approximately find the inlier overlap in constant time. 

After obtaining a set of instance clusters, the next step is to replace the instances in each cluster with a single one.
Even though it would be straightforward to use the density modes, \eg as in~\cite{comaniciu1999mean}, it requires doing operations with the preference vectors, \eg, averaging. However, such operations are undefined in the consensus space -- the average of two vectors is not necessarily the preference vector of the average instance.
Thus, we replace each cluster with one of its elements that has the highest quality $Q_f$ and, thus, is the most likely to represent the sought model parameters.

In the implementation, we use the DBSCAN~\cite{ester1996density,schubert2017dbscan} density-based clustering that runs swiftly on our problem and returns accurate solutions.
DBSCAN requires two parameters, \ie, the minimum size $c_\text{min}$ of a cluster to be kept and a threshold $\epsilon_\text{T}$ to decide if two model instances are neighbors in the consensus space.
The minimum size $c_\text{min} = 1$ since single-element clusters also contain dominant model instances and, thus, should be kept. 
The setting of threshold $\epsilon_\text{T}$ is intuitive. 
Setting $\epsilon_\text{T}$ to $0$ means that we consider models neighbors if and only if their preference vectors are exactly the same. 
Parameter $\epsilon_\text{T} = 1$ means that all methods are neighbors even if they do not share inliers.
%

\subsection{Improving Instance Parameters}

In order to improve the parameters of the instances kept by the clustering algorithm, we apply an iteratively re-weighted least-squares approach starting from the initial instance parameters.
We use the robust MAGSAC++ weights.

The model optimization and clustering are applied repeatedly since during the optimization step two instances might become similar and, thus, should be put in the same cluster. 
This iteration stops when only one-element clusters are returned by the applied clustering algorithm.

\subsection{Termination Criterion}

To decide when the algorithm should terminate, we use the criterion proposed in~\cite{barath2019progressive} that is 
%
    $n_i = (|\Points| - |\Instances|) \sqrt[m]{1 - \sqrt[k]{1 - \mu}} \leq m + 1$,
%
where $\mu$ is the required confidence in the results typically set to $0.95$ or $0.99$; $k$ is the number of iterations; $m$ is the size of the minimal sample; $n_i$ and $|\Points|$ are the number of inliers and points; and $|\Instances|$ is the cardinality of the united inlier sets of the kept model instances.
This criterion is triggered if the probability of having an unseen model with at least $m + 1$ inliers is smaller than $1 - \mu$.
Since we have a criterion for an instance being dominant, the upper bound $m + 1$ for $n_i$ can be replaced by $q_\text{min}$ to terminate when the probability of finding a dominant instance falls below $1 - \mu$.

\section{Connected Component Sampling}

There have been a number of algorithms proposed, \eg PROSAC~\cite{chum2005matching}, P-NAPSAC~\cite{barath2020magsac++}, to find samples that consist of data points stemming from the same model instance early.
When fitting multiple instances to real-world data, it usually is a reasonable assumption that the points form spatially coherent structures~\cite{nasuto2002napsac,isack2012energy,barath2018graph,barath2020magsac++}. 
We propose a deterministic sampling that returns the connected components in a progressively densified neighborhood-graph as samples. The algorithm is shown in Alg.~\ref{alg:ccsampler}.

\begin{algorithm}[t]
\begin{algorithmic}
\Require $r$, $r_\text{min}$, $r_\text{max}$, $n_{\text{steps}}$ -- current, min., max.\ neighborhood radius and partition number; \\ $\Points$ -- data points; $\mathcal{A}$ -- neighborhood-graph; $m$ -- sample size
\Ensure $\mathcal{S}$ -- sample
\If{$\neg$Initialized($\mathcal{A}$)}\Comment{Run only once}
    \State $\mathcal{A} \leftarrow $ BuildNeighborhood($\Points$, $r_\text{max}$)\Comment{Radius is $r_\text{max}$}
    \State{$r \leftarrow r_\text{min}$} \Comment{The max.\ radius in $\mathcal{A}$ for the next step}
    \State{$\mathcal{C} \leftarrow $ GetConnectedComponents($\mathcal{A}$, $r$)}
\EndIf
\While{Empty($\mathcal{C}$) $\wedge \; r \leq r_{\text{max}} $}
    \State{$r \leftarrow r + (r_\text{max} - r_\text{min}) / n_{\text{steps}}$} 
    \State{$\mathcal{C} \leftarrow $ GetConnectedComponents($\mathcal{A}$, $r$)}
\EndWhile
\State{$\mathcal{S} \leftarrow \emptyset$}
\If{$\neg$Empty($\mathcal{C}$)}
    \Repeat\Comment{Get the next largest \textit{dominant} instance }
        \State{$\mathcal{S} \leftarrow $ GetLargest($\mathcal{C}$), $\mathcal{C} \leftarrow \mathcal{C} \setminus \text{GetLargest}(\mathcal{C})$}
    \Until{$|\mathcal{S}| \geq m \vee \text{Empty}(\mathcal{C})$}
\EndIf
\If{$|\mathcal{S}| < m$}
    \State{$\mathcal{S} \leftarrow $ PROSAC($\mathcal{P}$, $m$)}
\EndIf
\end{algorithmic}
\caption{Connected Component Sampler: the next $\mathcal{S}$.}
\label{alg:ccsampler}
\end{algorithm}

The user-defined parameters are the minimum ($r_\text{min}$) and maximum ($r_\text{max}$) neighborhood radii and the number of steps when densifying the graph ($n_\text{steps}$).
As initialization, the method first builds neighborhood-graph $\mathcal{A}$ using the maximum radius. Then the connected components are selected from a sub-graph of $\mathcal{A}$ where all edges are ignored that are larger than the current radius $r$. This is done to avoid building $\mathcal{A}$ multiple times. 
The algorithm returns the largest connected component that has at least $m$ points. If there is no such component, it increases the neighborhood size by changing $r$. Note that the returned sample $\mathcal{S}$ is not necessarily a minimal sample.
If $r$ exceeds $r_\text{max}$, there are no reasonable structures and, thus, it starts sampling from all data points in a global manner by the PROSAC sampler.
Also note that while PROSAC is a safe-guard for cases where the data is not spatially coherent, it was \textit{never} 
executed in experiments of Sec. \ref{sec:results}.

\section{Experimental Results}
\label{sec:results}

\begin{figure}[t]
    \centering
    \includegraphics[trim=0 0 0 62,clip,width=0.95\columnwidth]{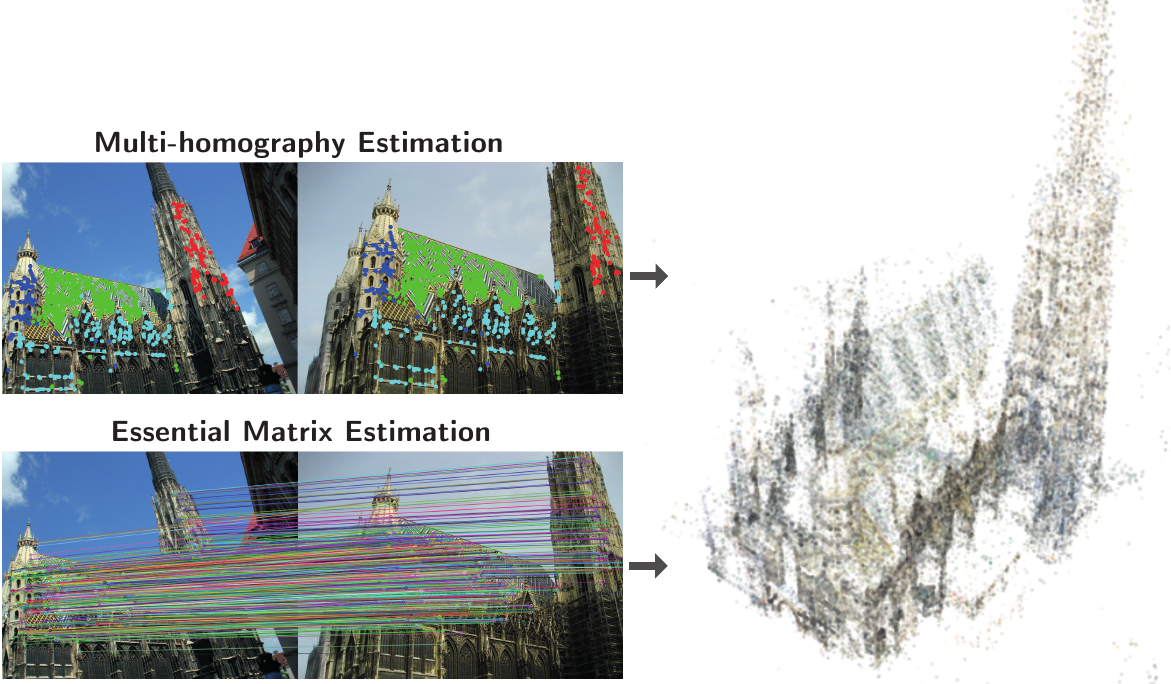}
	\caption{ Multiple \textbf{H}s contribute to the accurate reconstruction of Vienna Cathedral by \cite{sweeney2015theia}. The rot.\ and pos.\ errors decrease, respectively, by \SI{9.8}{\degree} and \SI{5.0}{\meter} compared to using \textbf{E} matrices only.}
	\label{fig:sfmvisualization}
\end{figure}


\customparagraph{Implementation Details.}
The \Method is implemented in C++ using the Eigen library and the solver implementation from the GC-RANSAC~\cite{barath2018graph} repository. 
We combine the algorithm with a number of components from USAC~\cite{raguram2013usac}.
The included components are the following.

\customsubparagraph{Sample degeneracy.} The degeneracy tests of minimal samples are for rejecting clearly bad samples to avoid the sometimes expensive model estimation. For homographies, samples consisting of collinear points are rejected.

\customsubparagraph{Sample cheirality.} The test is for rejecting samples based on the assumption that both cameras observing a 3D surface must be on its same side. For homography fitting, we check if the ordering of the four point correspondences (along their convex hulls) in both images are the same. 

\customsubparagraph{Model degeneracy.} The purpose of this test is to reject models early to avoid verifying them unnecessarily. For \textbf{F} matrices, DEGENSAC~\cite{chum2005two} is applied to determine whether the epipolar geometry is affected by a dominant plane. 


\customsubparagraph{Parameters.} Model-to-model threshold $\epsilon_\text{T} = 0.8$. 
This can roughly be interpreted as considering two model instances similar if more than $20\%$ of their inliers are shared.
The minimum quality $q_\text{min} = 20$ and confidence $\mu = 0.99$.
The parameters of the sampler are radii $r_\text{min} = 20$, $r_\text{max} = 200$, $r_\text{steps} = 5$. 
For point correspondences, the neighborhood is built on the joint 4D coordinate space. 
These parameters are used in \textit{all} tested problems and on all datasets. 
Additional explanation of the hyper-parameters is in the supp.\ material.

\begin{table*}[t]
	\center
	\resizebox{0.90\textwidth}{!}{
	\begin{tabular}{ r | c | c c c | c c c | c c c  }
    \hline
        \rowcolor{headergray} 
   		 & & \multicolumn{3}{ c | }{Adelaide: Two-view motions} & \multicolumn{3}{ c | }{Adelaide: Homographies} & \multicolumn{3}{ c }{Hopkins: Motions} \\
    \hline
   		 &  & \multicolumn{3}{ c | }{ 19 scenes } & \multicolumn{3}{ c | }{ 19 scenes } & \multicolumn{3}{ c  }{ 155 scenes }  \\
    \hline
   		 & $|\mathcal{I}|$ needed & \phantom{xx}avg.\phantom{xx} & \phantom{xx}std.\phantom{xx} & \phantom{xx}time\phantom{xx} & \phantom{xx}avg.\phantom{xx} & \phantom{xx}std.\phantom{xx} & \phantom{xx}time\phantom{xx} & \phantom{xx}avg.\phantom{xx} & \phantom{xx}std.\phantom{xx} & \phantom{xx}time\phantom{xx} \\
    \hline
        \rowcolor{mygray} 
   		\textbf{Proposed} & no & {\phantom{1}5.3} & \textbf{\phantom{1}4.4} & {\phantom{1}0.05} & \textbf{\phantom{1}3.1} & {\phantom{1}3.5} & \textbf{\phantom{11}0.11} & \textbf{\phantom{1}4.4} & \textbf{\phantom{1}6.3} & 0.04 \\
   		\rowcolor{mygray} 
   		\textbf{Proposed (CC)} & no & \textbf{\phantom{1}5.0} & {\phantom{1}\textbf{4.4}} & \textbf{\phantom{1}0.02} & {\phantom{1}5.7} & {\phantom{1}6.5} & \textbf{\phantom{11}0.11} & -- & -- & -- \\
   	    Prog-X \cite{barath2019progressive} & no & 10.7 &  {\phantom{1}8.7} & 14.38 & {\phantom{1}6.6} & {\phantom{1}5.9} & \phantom{11}1.03 & {\phantom{1}8.4} & 10.3 & \textbf{0.02}  \\
   		Multi-X \cite{barath2018multix} & no & 17.1 & 12.2 & \phantom{1}{1.52} & \phantom{1}8.7 & \phantom{1}8.1 &  \phantom{11}{0.27} & 13.0 & 19.6 & 0.95 \\ 
   		PEARL~\cite{isack2012energy} & no & 29.5 & 14.8 & \phantom{1}4.94 & 15.1 & \phantom{1}6.8 & \phantom{11}2.61 & 14.3 & 23.2 & 3.30 \\ 
   		RPA~\cite{magri2015robust} & yes & 17.1 & 11.1 & 10.24 & 23.5 & 13.4 & 622.87 & \phantom{1}9.2 & 11.3 & 4.92 \\ 
   		RansaCov~\cite{magri2016multiple} & yes & 55.6 & 12.4 & \phantom{1}2.33 & 66.9 & 18.4 & \phantom{1}17.69 & 11.1 & \phantom{1}{8.0} & 2.04 \\
   	    T-linkage~\cite{magri2014t} & no & 46.7 & 15.6 & \phantom{1}2.69 & 54.8 & 22.2 & \phantom{1}57.84 & 27.2 & 15.6 & 0.95  \\
   		MLink~\cite{magri2021multilink} & no & \phantom{1}8.6 & \phantom{1}4.7 & 16.75 & {\phantom{1}5.5} & \phantom{1}\textbf{1.8} & \phantom{1}47.75 & \phantom{1}8.3 & 11.9 & -- \\
   		CONSAC~\cite{kluger2020consac} & no & -- & -- & -- & {\phantom{1}5.2} & \phantom{1}6.5 & \footnotesize 8.1 / 21.0 & -- & -- & -- \\  
    \hline
    \hline      
\end{tabular}}\vspace{1mm}
\caption{Avg.\ misclassification errors (in \%; 5 runs on each scene), their std.\ and the run-times (secs) on two-view motion and homography fitting on the AdelaideRMF dataset~\cite{wongiccv2011}, and motion fitting on the Hopkins dataset~\cite{tron2007benchmark}. 
All methods use fixed parameters.
For CONSAC, we report the times of running it on GPU and CPU. 
The second column ($|\mathcal{I}|$ needed) is ``yes'' for methods requiring the number of instances to fit. 
In the first row, the proposed method runs the P-NAPSAC~\cite{barath2020magsac++} sampler.
In the second one, the proposed CC sampler is used. }
\label{tab:summary_table}
\end{table*}

\subsection{Standard Benchmarks}

To evaluate the proposed method on real-world problems, we use a number of publicly available datasets for homography, two-view motion, and motion fitting. 
The error is the misclassification error (ME), \ie, the ratio of points assigned to the wrong cluster.
%
%
The \Method is designed to avoid assigning each data point to a single instance. 
Thus, we assigned each point to the model with the smallest residual.
The results of the compared methods are copied from~\cite{barath2019progressive,kluger2020consac,magri2021multilink}, where they were carefully tuned to achieve their best results with fixed parameters.

To test the proposed connected component-based sampler, we applied the \Method with P-NAPSAC~\cite{barath2020magsac++} and the proposed Connected Component Sampler, both of them exploiting the spatial nature of geometric data.
We chose P-NAPSAC as a competitor, since it has a similar procedure, finding local structures by randomly sampling from gradually growing neighborhoods.
The major difference between them is that P-NAPSAC is randomized and returns minimal samples, while the proposed Connected Component Sampler is deterministic and proposes larger-than-minimal samples as well.

Examples of multi-homography and two-view motion fitting are in Fig.~\ref{fig:example_results}. 
We chose the scenes from the AdelaideRMF~\cite{wongiccv2011} dataset with the most ground truth models to be found. 
Color denotes the point-to-model assignment done by 
assigning each point to the instance, outputted by the proposed method,  
with the smallest residual. 

\customparagraph{Two-view motion} fitting is tested on the AdelaideRMF motion dataset consisting of $19$ image pairs and correspondences manually assigned to two-view motion clusters. 
In this case, multiple \textbf{F} matrices are to be found. 
For the proposal step, we used the 7PT algorithm~\cite{hartley2003multiple}.
In the IRLS fitting, we applied the norm.\ 8PT solver~\cite{hartley1997defense}.

The avg.\ errors over five runs and their std.\ are shown in the left block of Table~\ref{tab:summary_table}. 
The proposed method leads to state-of-the-art accuracy with both tested samplers. 
The proposed Connected Components Sampler (CC) improves both the accuracy and processing time.
The \Method with CC is twice as accurate as the second best competitor (MLink) while being two orders-of-magnitude faster than the second fastest method (Multi-X).
The \Method runs in \textit{real-time} on these scenes. 
%
On avg., out of the $45$ motions in the dataset, the proposed method does not find $2$ instances while returning $1$ false positive.

\customparagraph{Homography} fitting is tested on the AdelaideRMF H dataset~\cite{wongiccv2011}. 
It consists of $19$ image pairs with ground truth correspondences assigned manually to \textbf{H}s. 
In these tests, we also included the errors of CONSAC~\cite{kluger2020consac}. 
Since the run-times are not reported in~\cite{kluger2020consac}, we re-ran the algorithm both on GPU and CPU and calculated the avg.\ times.  

We used the norm.\ 4PT algorithm both in the proposal and IRLS steps.
The results are shown in the middle block of Table~\ref{tab:summary_table}. 
The \Method is almost twice as accurate as the second best one (CONSAC) while being significantly faster than all algorithms. 
It leads to the most accurate solutions while being the fastest. 
In this case, P-NAPSAC sampler leads to the best results.
Out of the 52 \textbf{H}s, the proposed method does not find $2$ with $2$ false positives.

\customparagraph{Motion} segmentation is tested on $155$ videos of the Hopkins dataset~\cite{tron2007benchmark}. 
It consists of $155$ sequences divided into three categories: {\fontfamily{cmtt}\selectfont{checkerboard}}, {\fontfamily{cmtt}\selectfont{other}}, and {\fontfamily{cmtt}\selectfont{traffic}}. 
The trajectories are inherently corrupted by noise, but no outliers are present.
Motion segmentation in videos is the retrieval of sets of points undergoing rigid motions in a dynamic scene captured by a moving camera.
It can be considered a subspace segmentation under the assumption of affine cameras. 
For such cameras, all feature trajectories associated with a single moving object lie in a 4D linear subspace in $\mathbb{R}^{2F}$, where $F$ is the frame number~\cite{tron2007benchmark}.

The results are shown in the right part of Table~\ref{tab:summary_table}. 
The proposed method leads to the lowest errors. 
It still runs in \textit{real-time}. 
In this case, we used uniform sampling since building a neighborhood-graph (required both by the CC sampler and P-NAPSAC) on point trajectories is not trivial.   



\begin{table}[t]
    \centering
    \begin{minipage}[t][][b]{0.49\textwidth}
        \centering
        \resizebox{0.88\columnwidth}{!}{\begin{tabular}{ r | c c | c c  }
            \hline
                \rowcolor{headergray}
                \multicolumn{5}{ c }{ Relative Pose Estimation } \\
            \hline
           		&  avg. $\epsilon_\mathbf{R}$ &  med. $\epsilon_\mathbf{R}$	&  avg. $\epsilon_\mathbf{t}$ &  med. $\epsilon_\mathbf{t}$  \\    
            \hline
                \textbf{E} matrix & 9.51 & 3.46 & 18.15 & \phantom{2}9.08 \\        \rowcolor{mygray} 
                \textbf{E} from \textbf{H}s & 9.56 & 3.47 & 18.21 & \phantom{2}9.09 \\  \rowcolor{mygray}
                Pose averaging & 8.71 & 3.69 & 34.34 & 25.27 \\  \rowcolor{mygray}
                Pose selection & {8.33} & {3.34} & {17.84} & \phantom{2}{8.92} \\
                \rowcolor{mygray}
                Pose selection (CC) & \textbf{8.24} & \textbf{3.31} & \textbf{17.81} & \phantom{2}\textbf{8.89} \\
            \hline
            \hline
        \end{tabular}}
        \caption{
            Relative rotation $\epsilon_\textbf{R}$ and translation $\epsilon_\textbf{t}$ errors ($^\circ$) on \num{435}k image pairs from the 1DSfM dataset obtained by \textbf{E} matrix estimation; calculating \textbf{E} from the inliers of homographies (\textbf{E} from \textbf{H}s); pose averaging on the poses decomposed from \textbf{E} and multiple \textbf{H}s; and selecting the pose with the most inliers from the decomposed ones (Pose selection) with the proposed sampler (CC).
            }
        \label{tab:pose_table_rel}
    \end{minipage}\hfill
    \begin{minipage}[t][][b]{0.49\textwidth}
        \centering
        \resizebox{0.88\columnwidth}{!}{\begin{tabular}{ r | c c | c c  }
            \hline
                 \rowcolor{headergray}
                \multicolumn{5}{ c }{ Global SfM Results } \\
            \hline
           		&  avg. $\epsilon_\mathbf{R}$ &  med. $\epsilon_\mathbf{R}$	&  avg. $\epsilon_\mathbf{p}$ &  med. $\epsilon_\mathbf{p}$  \\  
            \hline  
                \textbf{E} matrix & 11.15 & 6.58 & {10.25} & 8.93 \\      \rowcolor{mygray}
                \textbf{E} + mult. \textbf{H}s & \phantom{1}{7.93} & {6.21} & 10.52 & {4.60}  \\  \rowcolor{mygray}
                \textbf{E} + mult. \textbf{H}s (CC) & \phantom{1}\textbf{5.56} & \textbf{5.61} & \phantom{1}\textbf{9.57} & \textbf{3.99}  \\ 
            \hline
            \hline
        \end{tabular}}
        \caption{ Rotation and position errors of the global SfM implemented in~\cite{sweeney2015theia} when initialized with a poses estimated from \textbf{E} matrices, and via the proposed pose selection from \textbf{E} and multiple \textbf{H}s.  }
        \label{tab:pose_table_global}
    \end{minipage}
\end{table}

\subsection{Application: Relative Pose Estimation}
\label{sec:exp_relative_pose}

In this section, we focus on improving relative pose estimation by exploiting multiple homographies. 
Pose estimation is a fundamental problem in a number of popular methods, \eg, in Structure-from-Motion algorithms. 
While the usual procedure to estimate a relative pose uses epipolar geometry, it is well-known that the pose can also be obtained from a homography if the cameras are calibrated. 
However, in most pipelines, homographies are used only if the scene is degenerate for fundamental matrix estimation, \eg, a single plane dominates the scene~\cite{chum2005two} or the camera undergoes purely rotational motion.
In this section, we aim to propose a way of exploiting multiple homographies to improve the relative pose accuracy. 
See Fig.~\ref{fig:sfmvisualization} for an example.

We downloaded the 1DSfM dataset~\cite{wilson_eccv2014_1dsfm} and applied COLMAP~\cite{schonberger2016structure} to obtain a reconstruction that can be used as ground truth. 
Note that the 1DSfM dataset provides a ground truth, however, it was created by the Bundler algorithm~\cite{snavely2006photo} that is more than 10 years old.
We use the following approach in order to find potentially matching image pairs. 
First, we extract GeM~\cite{GeM2018} descriptors with ResNet-50~\cite{He2016ResNet} CNN, pre-trained on GLD-v1 dataset~\cite{DELF2017}. Then we calculate the inner-product similarity between the descriptors, resulting in an $n\times n$ similarity matrix. In the experiments, we use only the image pairs with similarity higher than $0.4$ \cite{barath2021posegraph}. 
Finally, we estimated multiple homographies for all considered image pairs, $\num{434587}$ in total.

We tested the following approaches to recover the relative pose from multiple \textbf{H}s: 
\begin{enumerate}[topsep=1pt, partopsep=1pt,itemsep=1pt,parsep=1pt]
\item Estimating the essential matrix~\cite{stewenius2006recent} from the inliers of the returned homographies.
\item Decomposing all found homographies and, also, the essential matrix relative poses and running pose averaging by~\cite{chatterjee2013efficient,wilson2014robust}.
\item Decomposing each homography~\cite{malis2007deeper} and, also, the essential matrix to pose and selecting the one which has the most inliers determined by thresholding the re-projection error. The translation is then re-estimated by solving equation $\textbf{p}_2^\text{T} [\textbf{t}]_\times \textbf{R} \textbf{p}_1 = 0$ with known rotation \textbf{R}, where $[\textbf{t}]_\times$ is the cross-product matrix of translation $\textbf{t}$ and $[\textbf{t}]_\times \textbf{R}$ is the essential matrix. Details are in the supplementary material.
\end{enumerate}
\noindent
The results are reported in Table~\ref{tab:pose_table_rel}. 
Results when using the proposed connected component sampler (CC) are also shown.
To measure the error in the rotation, we calculate the angular difference between the ground truth $\dot{ \mathbf R}$ and estimated $\mathbf R$ ones as $\epsilon_\textbf{R} = \cos^{-1}((\text{tr}(\mathbf R  \dot{ \mathbf R}^\trans)-1)/2)$. Since the translation is up to scale, the error is the angular difference $\epsilon_\textbf{t}$ of the ground truth and estimated translations. 
The avg.\ rotation and translation errors are improved by, respectively, $1.27$ and $0.34$ degrees compared to \textbf{E} estimation.
CC sampler leads to the best results.
Since it is extremely fast, the computational overhead is merely a few \textit{ms}. 

We applied the global SfM implemented in the Theia library~\cite{sweeney2015theia} initialized with the poses estimated in the proposed way and, also, with the poses estimated using only essential matrices.
The accuracy of the reconstruction is reported in Table~\ref{tab:pose_table_global}. 
We report the average rotation (avg.\ $\epsilon_{\textbf{R}}$, in degrees) and position errors (avg.\ $\epsilon_{\textbf{p}}$, in meters) and, also, the median errors averaged over the scenes. 
The proposed algorithm with the CC sampler \textit{significantly} reduces both the rotation and position errors of the reconstruction.


\begin{table}
	\center
	\resizebox{0.88\columnwidth}{!}{
	\begin{tabular}{ r | c c | c c  }
    \hline \rowcolor{headergray} 
   		&  avg. $\epsilon_\mathbf{R}$ &  med. $\epsilon_\mathbf{R}$	&  avg. $\epsilon_\mathbf{p}$ &  med. $\epsilon_\mathbf{p}$  \\                                  
    \hline
        \textbf{E4+2} & 1.19 & 0.50 & 0.033 & 0.025 \\       
        \textbf{H3+2} & 0.45 & \textbf{0.24} & 0.103 & 0.041 \\  \rowcolor{mygray}      
        \textbf{E} + mult. \textbf{H}s & \textbf{0.32} & 0.26 & \textbf{0.026} & \textbf{0.022} \\
    \hline
    \hline
\end{tabular}}
\caption{ The avg.\ and med.\ rotation $\epsilon_\textbf{R}$ ($^\circ$) and position $\epsilon_\textbf{p}$ (\textit{m}) errors on \num{23190} image pairs from the KITTI dataset obtained by generalized \textbf{E} matrix (\textbf{E4+2}) \cite{zheng2015structure} and \textbf{H} estimation~\cite{bhayani2021calibrated} (\textbf{H3+2}); and by selecting the pose obtained from a generalized \textbf{E} matrix and a set of generalized homographies (\textbf{E} + mult. \textbf{H}s). }
\label{tab:generalized_pose_table}
\end{table}

\subsection{Application: Fast-moving Object Detection}

In this section, we estimate the trajectories of objects that are significantly blurred by their motion.
As defined in~\cite{fmo}, an image $I$ of such blurred object is formed as a composition of the blurred object appearance and the background
\begin{equation}
	\label{eq:blatting}
	I = H*F + (1-H*M)\,B,
\end{equation}
where the sharp object appearance $F$ with mask $M$ encodes the object, blur kernel $H$ encodes the trajectory, and $B$ represents the background. 
Input image $I$ and background $B$ are assumed to be known. 
The unknowns in~\eqref{eq:blatting} are estimated either by alternating energy minimization with additional priors~\cite{tbd,tbd3d,tbd_ijcv,tbdnc,kotera2018,kotera2020} or more recently by learning from synthetic data~\cite{defmo,fmodetect} and neural rendering~\cite{sfb,mfb}.

\begin{figure}
\centering
\setlength{\tabcolsep}{0.05em} 
\renewcommand{\arraystretch}{0.25} 
\begin{tabular}{@{}cccc@{}}
\includegraphics[trim=150 0 150 0,clip,width=0.245\columnwidth]{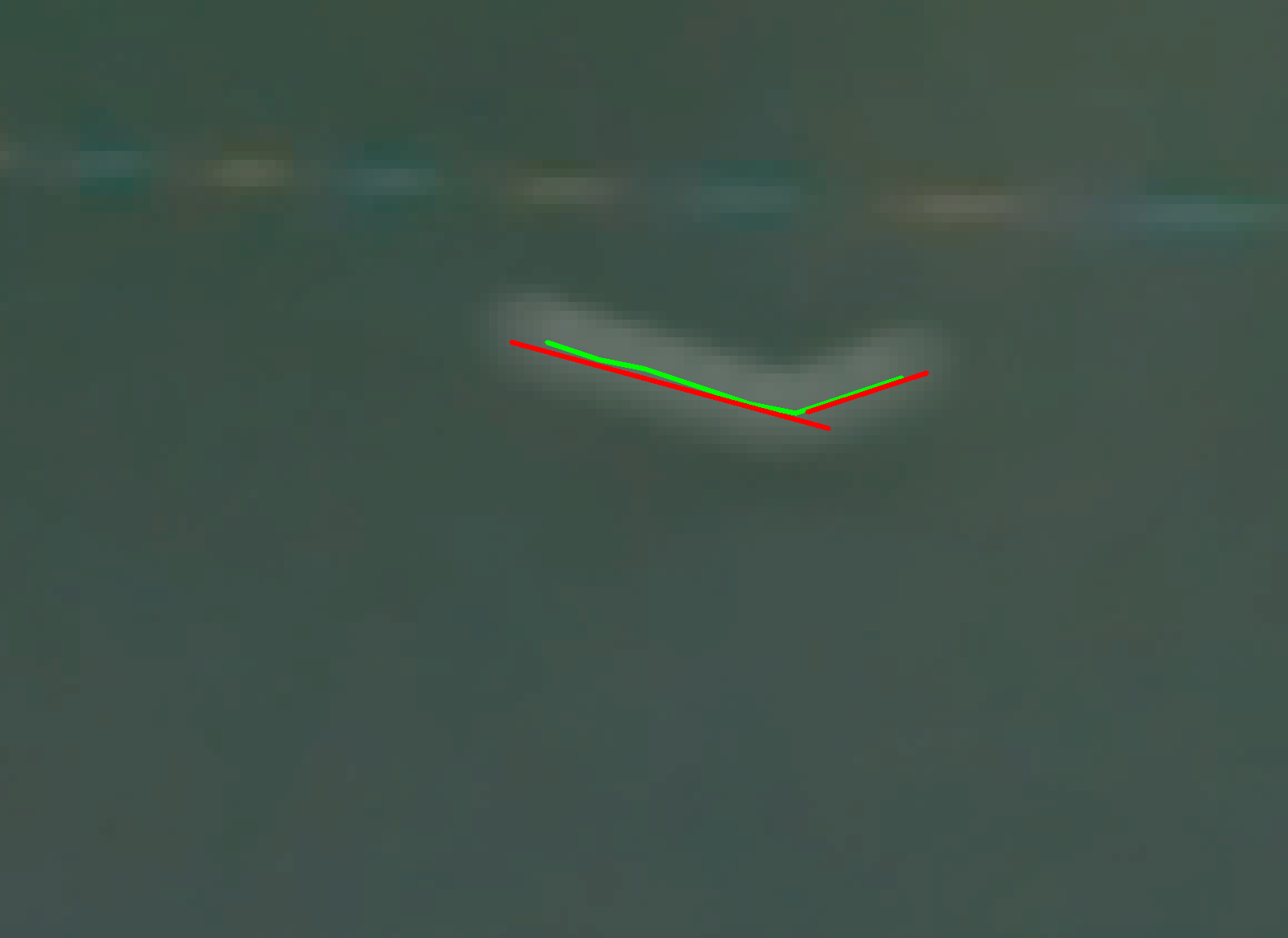}\llap{\makebox[0.245\columnwidth][l]{\includegraphics[trim=7 7 6 6,clip,height=0.6cm]{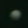}}} &
\includegraphics[trim=150 0 150 0,clip,width=0.245\columnwidth]{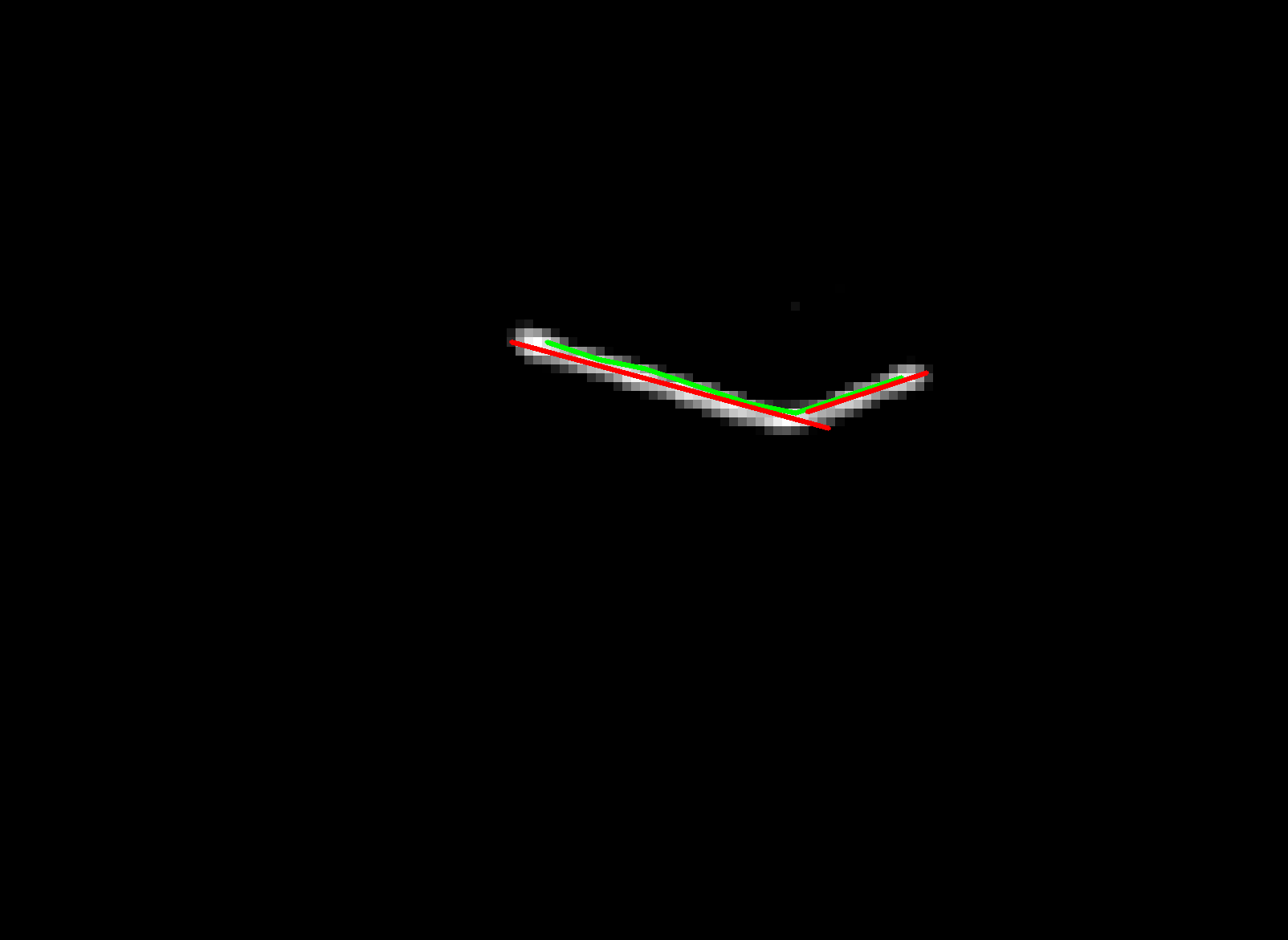} &
\includegraphics[trim=150 185 150 0,clip,width=0.245\columnwidth]{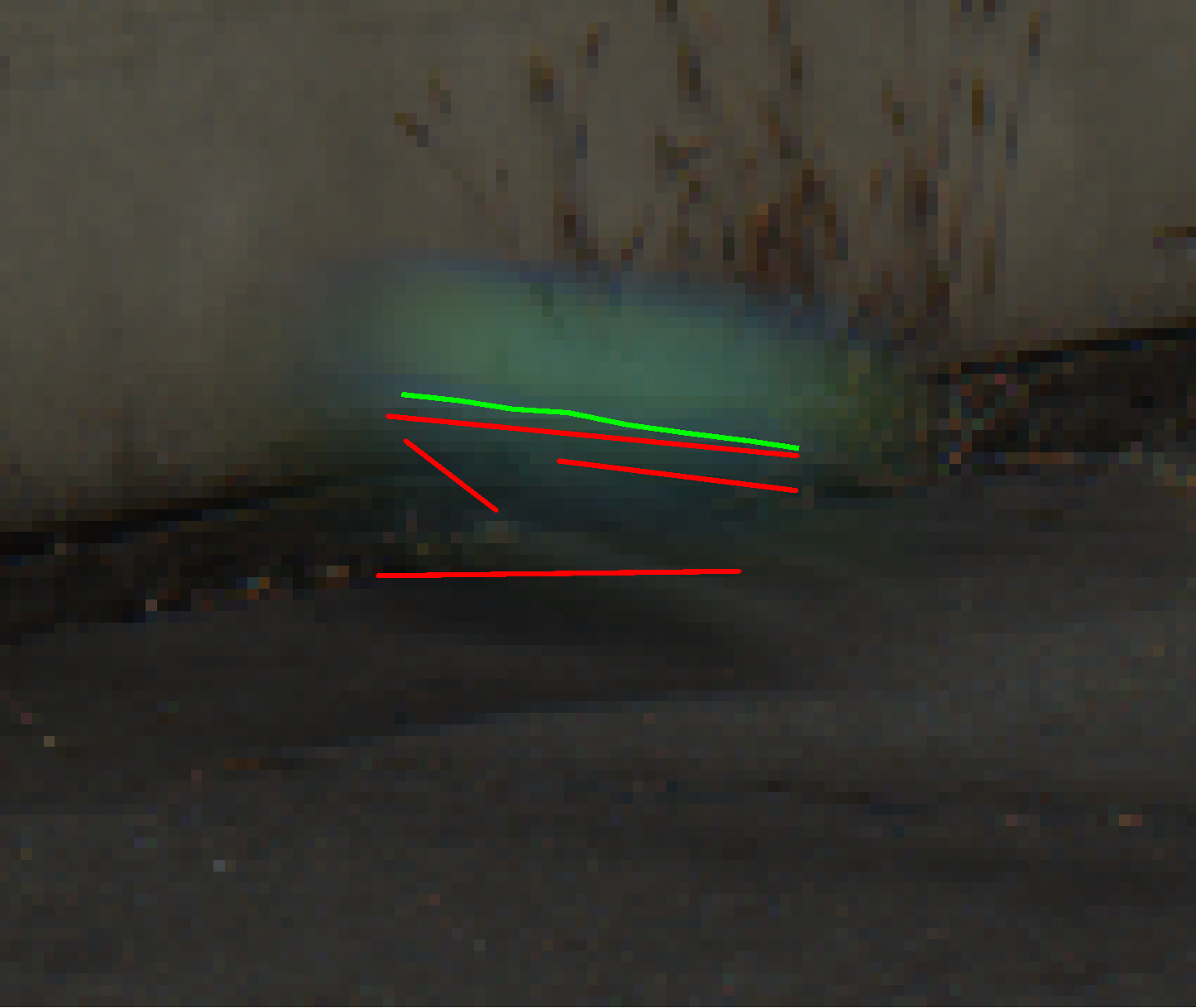}\llap{\makebox[0.245\columnwidth][l]{\includegraphics[height=0.6cm]{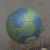}}} &
\includegraphics[trim=150 180 150 0,clip,width=0.245\columnwidth]{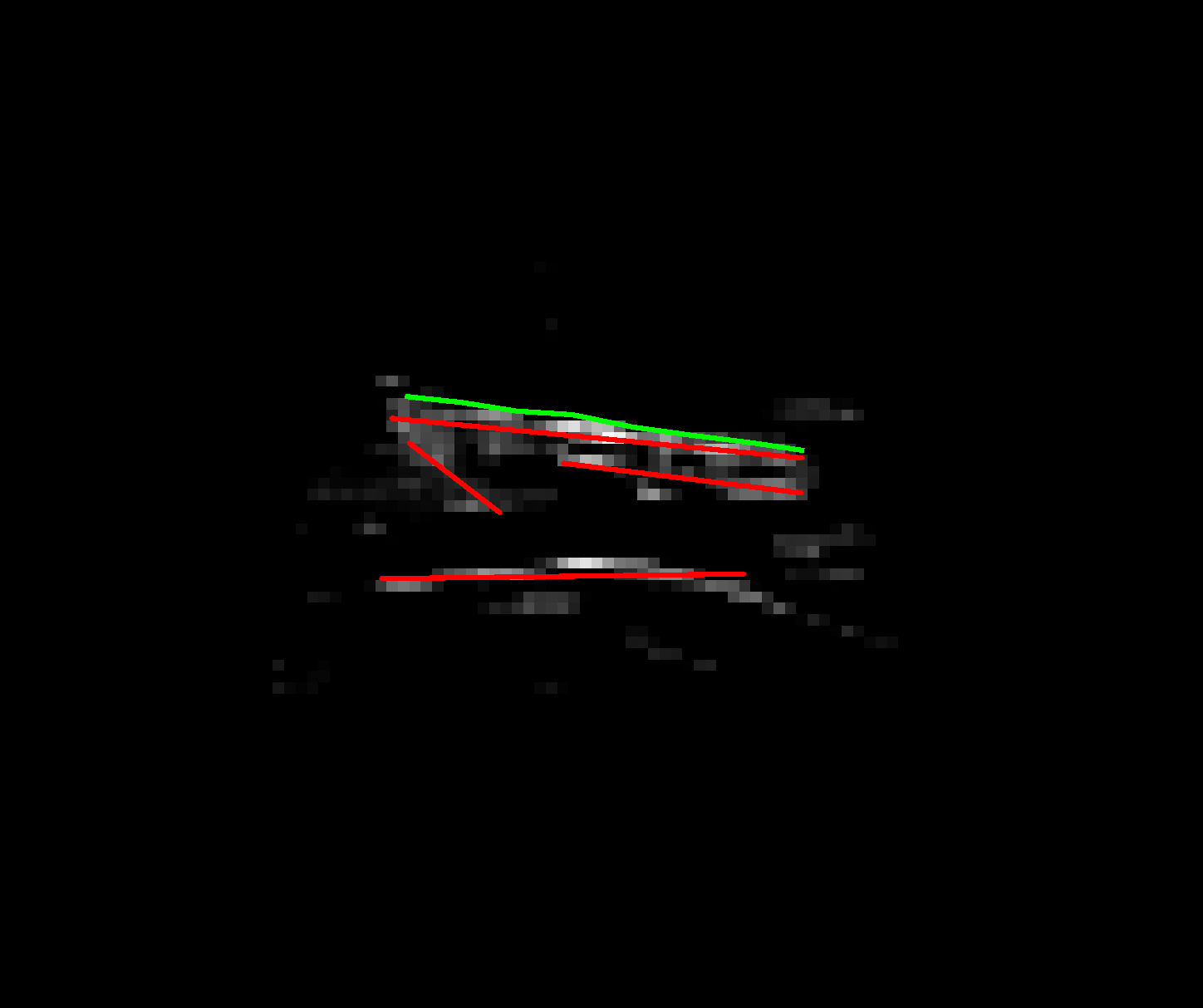}  \\[0.1em]
Input image & Blur kernel & Input image & Blur kernel \\
\end{tabular}
\caption{Multiple line segment fitting for trajectory estimation of fast-moving objects. Estimated line segments are in red, the ground truth is in green. Sharp object appearance is overlaid in the bottom left corner of the input image. }
\label{fig:fmo}
\end{figure}

The formation model in~\eqref{eq:blatting} encodes the trajectory by the blur kernel. 
However, there are no guarantees that the blur kernel corresponds to a physically plausible trajectory, which is assumed to be piece-wise linear due to bounces.
Blur kernels also contain other responses due to other moving objects in the scene.
In the extreme case, if two fast-moving objects intersect or fly close to each other, the blur kernel will contain multiple responses corresponding to each motion.
In practice, the estimated blur kernels are noisy, with many outliers, and contain artifacts due to shadows, low contrast, and discretization.
Motion blur priors~\cite{sroubek2020} have been proposed to reduce these issues, but extracting the final continuous trajectory is still a challenging multi-instance model fitting task (see Fig.~\ref{fig:fmo} for examples).

Recent methods~\cite{tbd,tbd3d} address this task by employing Sequential RANSAC~\cite{vincent2001detecting,kanazawa2004detection} on the thresholded blur kernels.
We extract blur kernels using the TbD method~\cite{tbd} from all sequences in the TbD~\cite{tbd} and TbD-3D~\cite{tbd3d} datasets. 
The TbD dataset is simpler since it contains mostly uniformly colored objects moving in the plane parallel to the camera plane.
The TbD-3D dataset is more challenging with highly textured objects that are rotating and moving in 3D.
The ground truth sub-frame object location is given from a high-speed camera.
We estimate multiple line segments in each blur kernel and measure the average $L_2$ distance of each ground truth location to the closest fitted line segment.
Table~\ref{tab:line_segment_table} shows the average error, its standard deviation, and average run-time for a wide range of state-of-the-art methods. 
We used the implementations provided by the authors. 
The proposed method outperforms all compared algorithms both in terms of accuracy and processing time, running in \textit{real-time}.
Additional results, \eg demonstrating the effect of the proposed soft assignment, are in the supplementary material.
Without considering soft assignment, continuous chains can not be found. This leads to losing short segments and affects the accuracy notably.

\begin{table}
	\center
	\resizebox{0.88\columnwidth}{!}{
	\begin{tabular}{ r | c c c | c c c  }
    \hline
    \rowcolor{headergray} 
	Dataset:  & \multicolumn{3}{ c | }{  \textit{Easy} (322)~\cite{tbd} }  & \multicolumn{3}{ c  }{  \textit{Challenging} (470)~\cite{tbd3d} } \\
    \hline
    \hline
   		&  avg. &  std. &  time &  avg. &  std. &  time  \\     
    \hline
    \rowcolor{mygray} 
        \textbf{Proposed} & \textbf{1.39} & 6.73 & \textbf{0.02} & \textbf{2.84} & \textbf{2.80} & \phantom{1}\textbf{0.05} \\
        {Prog-X~\cite{barath2019progressive}} & 1.87 & 6.80 & 0.24 & 3.74 & 3.22 & \phantom{1}0.09 \\
        {PEARL~\cite{isack2012energy}} & \textbf{1.39} & 6.74 & 0.05 & 4.83 & 6.17 & \phantom{1}0.08 \\ 
        {J-Linkage~\cite{toldo2008robust}} & 1.73 & 6.72 & 4.02 & 4.85 & 6.51 & \phantom{1}4.52 \\ 
        {T-Linkage~\cite{magri2014t}} & 1.71 & \textbf{6.71} & 7.07 & 4.46 & 5.21 & 33.65 \\ 
        {RPA~\cite{magri2015robust}} & 2.74 & 7.77 & 7.66 & 5.19  &  4.47 & 21.79  \\ 
        {RansaCov~\cite{magri2016multiple}}  & 1.48 & 6.74 & 2.09 & 3.90  & 4.83  & \phantom{1}7.62  \\ 
        {Seq. RANSAC} & 1.66 & 6.72 & 0.68 & 6.08 & 7.50 & \phantom{1}0.98 \\
    \hline
    \hline
\end{tabular}}
\caption{The avg.\ and std.\ accuracy (px) and run-time (secs) of multiple line segment detection for finding the trajectories of fast-moving objects. The number of images are in brackets. }
\label{tab:line_segment_table}
\end{table}

\subsection{Pose from Generalized Camera}

To further test the pose selection technique from multiple homographies and an essential matrix, as proposed in Section~\ref{sec:exp_relative_pose}, we downloaded the KITTI odometry dataset~\cite{Geiger2012CVPR}, where each frame consists of the images of two cameras. 
We considered the two cameras as a generalized one and estimated the pose between this camera and the left image of the next frame. 
We used the generalized essential matrix~\cite{zheng2015structure} (\textbf{E4+2}) and homography~\cite{bhayani2021calibrated} (\textbf{H3+2}) solvers. %
For finding a single \textbf{E4+2} or \textbf{H3+2}, we used GC-RANSAC~\cite{barath2018graph}.
The methods were tested on a total of \num{23190} frame pairs. 

The results are in Table~\ref{tab:generalized_pose_table}. 
The proposed technique (\textbf{E} + mult.\ \textbf{H}s), selecting the best pose from the set decomposed from an essential matrix and multiple homographies, 
leads to the most accurate results in terms of average rotation and position errors. Its median rotation error is similar to $\textbf{H3+2}$. Its median position error is the lowest.

\blfootnote{\scriptsize \textbf{Acknowledgements.} This project was supported by the ETH Postdoc fellowship, by the OP VVV funded project CZ.02.1.01/0.0/0.0/16$\_$019/0000765 ``Research Center for Informatics'' (RCI), and by the Robert Bosch GmbH. The conference presentation was supported by Czech Technical University in Prague (RCI project).}

\section{Conclusion}

We propose a new multi-instance model fitting algorithm that is a simple iteration of instance proposal, clustering in the consensus space, and parameter re-estimation.
Due to not forming crisp point-to-model assignments, the method runs in \textit{real-time} on a number of vision problems.
On two-view motion estimation, it is at least two orders-of-magnitude faster than the competitors.
It leads to results superior to the state-of-the-art both in terms of accuracy and run-time on the standard benchmark datasets.  
Moreover, the proposed Connected Component sampler outperforms the recent P-NAPSAC on a number of real-world problems.
In addition, we demonstrated on a total of \num{458569} images or image pairs that using multiple model instances, \eg homographies or line segments, is beneficial for various popular vision applications, \eg, Structure-from-Motion.

\noindent

\appendix
\section{Explanation of the Hyper-parameters}

In this section, we describe the hyper-parameters of the proposed algorithm, their purpose and the ways to set them.
Parameters of the proposed algorithm:
\begin{enumerate}
    \item An upper-bound for the inlier-outlier threshold on the point-to-model residual used inside the MAGSAC++ scoring. This parameter is problem-dependent. It usually is defined in pixels. It is easier to set~\cite{jin2021image} than the usual inlier-outlier threshold of RANSAC. 
    \item Parameter $q_{\text{min}}$ is similar to what structure-from-motion algorithms use to decide if the relative pose of an image pair is estimated successfully. For example, COLMAP~\cite{schonberger2016structure} uses $q_{\text{min}}$ = 15, we use 20.
    \item The termination confidence is the same as in RANSAC. Its typical values are 0.95 and 0.99. We use 0.99 in our experiments.
    \item The model-to-model distance threshold is from interval $\in[0,1]$. It measures the overlap of the inlier sets of two models (0 - non-overlapping, 1 - fully overlapping). Setting it to 0.2 works on a wide range of problems and datasets. 
\end{enumerate}

\section{SfM Results in Section 4.2}

\noindent
\textbf{Detailed results.}
The results of the global SfM from~\cite{sweeney2015theia} on each scene from the 1DSfM dataset are reported in Table~\ref{tab:1DSfM}.
Note that we omitted the results on scenes Gendarmenmarkt and Union Square since \cite{sweeney2015theia} failed to reconstruct them with all tested pose-graph estimation techniques.

\begin{figure*}
    \centering
    \begin{subfigure}[t]{0.48\textwidth}
   	 	\centering
        \includegraphics[width=1.0\columnwidth]{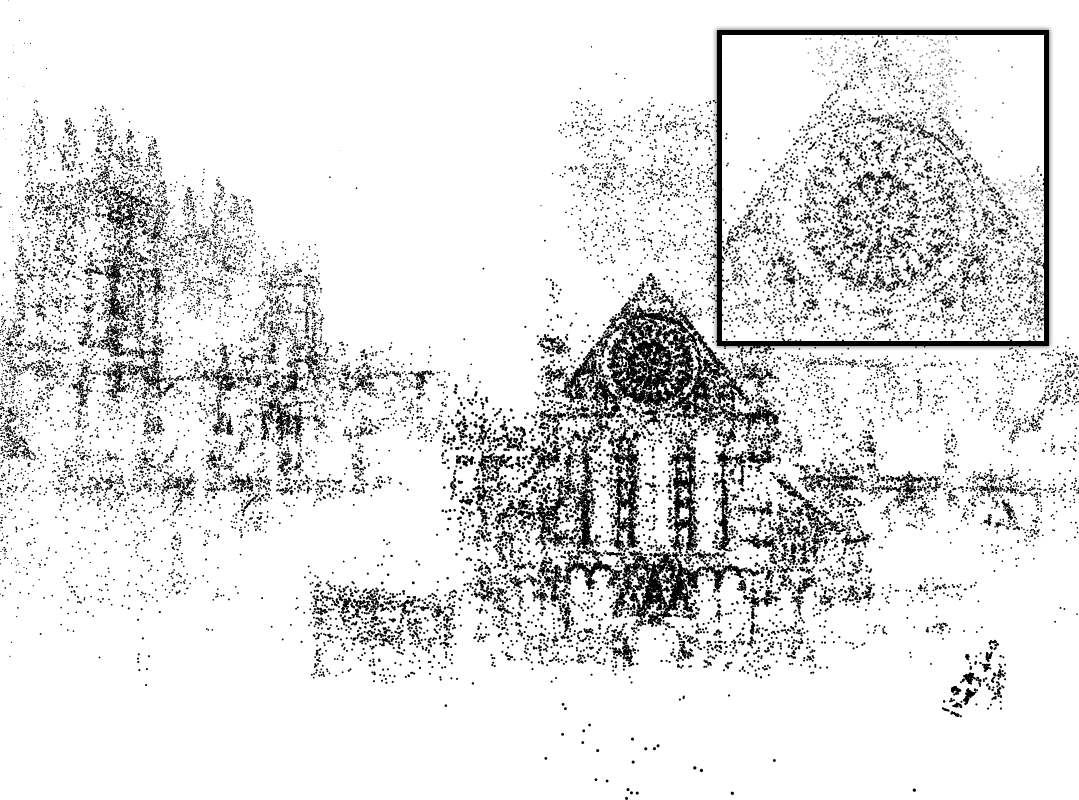}
        \caption{Frontal view -- \textbf{E} + mult. \textbf{H}s (CC).}
    \end{subfigure}\hfill
    \begin{subfigure}[t]{0.48\textwidth}
   	 	\centering
        \includegraphics[width=1.0\columnwidth]{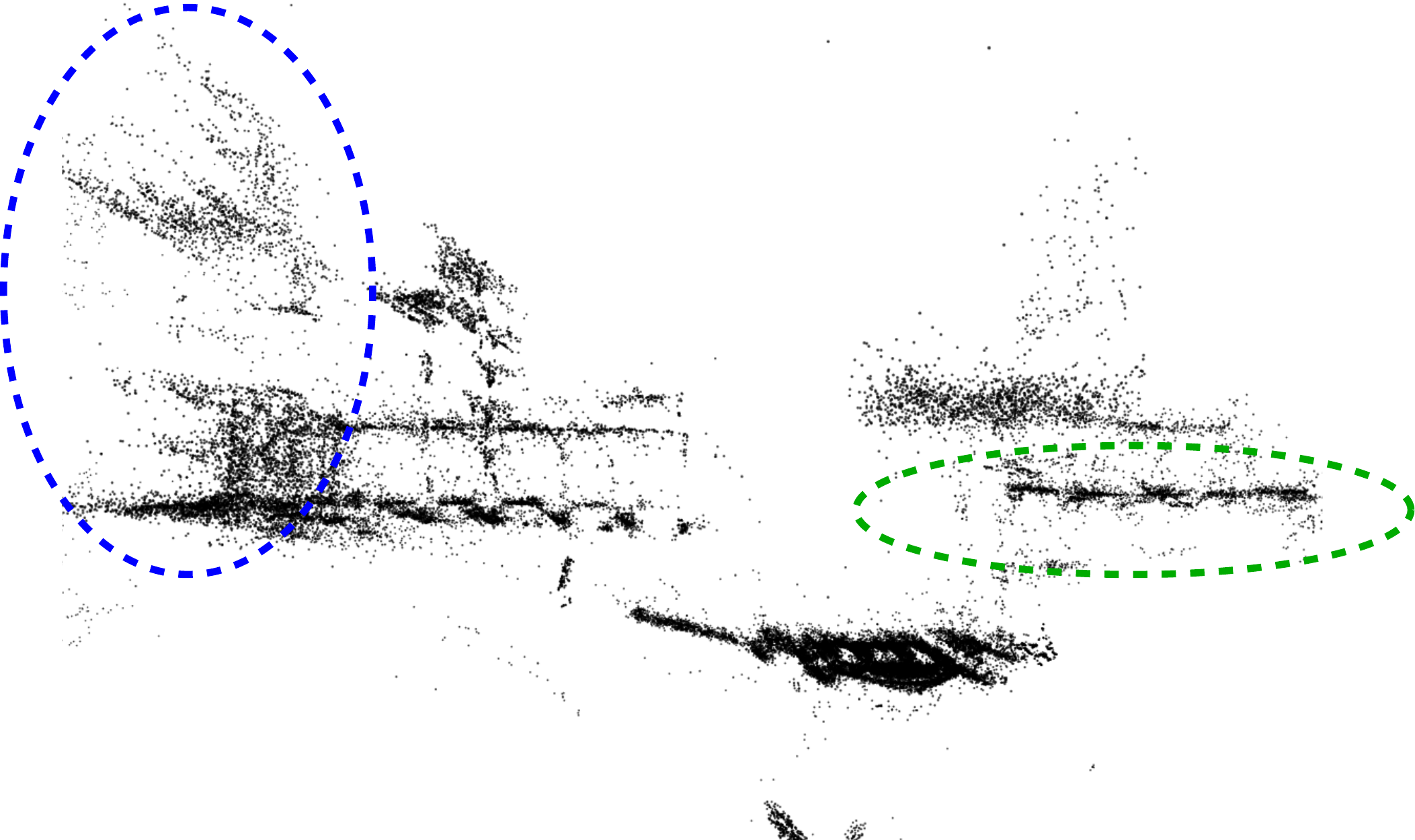}
        \caption{Top-down view -- \textbf{E} + mult. \textbf{H}s (CC).}
    \end{subfigure}
    \begin{subfigure}[t]{0.48\textwidth}
   	 	\centering
        \includegraphics[width=1.0\columnwidth]{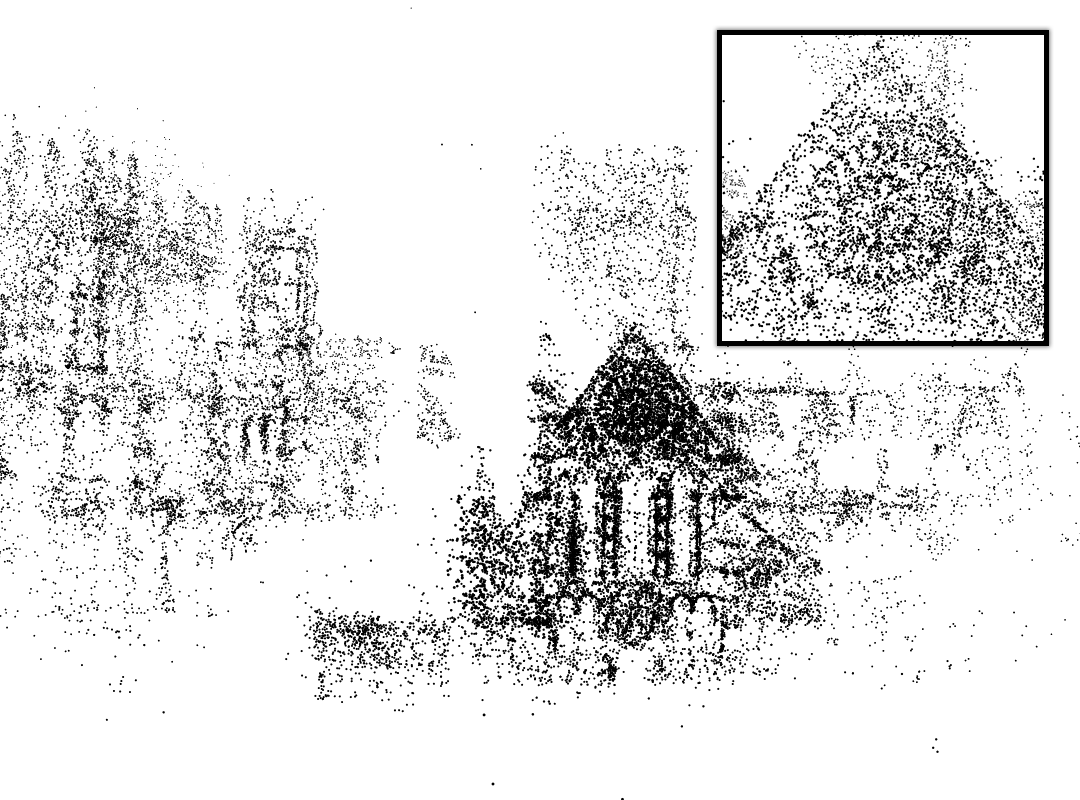}\\
        \caption{Frontal view -- \textbf{E} matrices only.}
    \end{subfigure}\hfill
    \begin{subfigure}[t]{0.48\textwidth}
   	 	\centering
        \includegraphics[width=1.0\columnwidth]{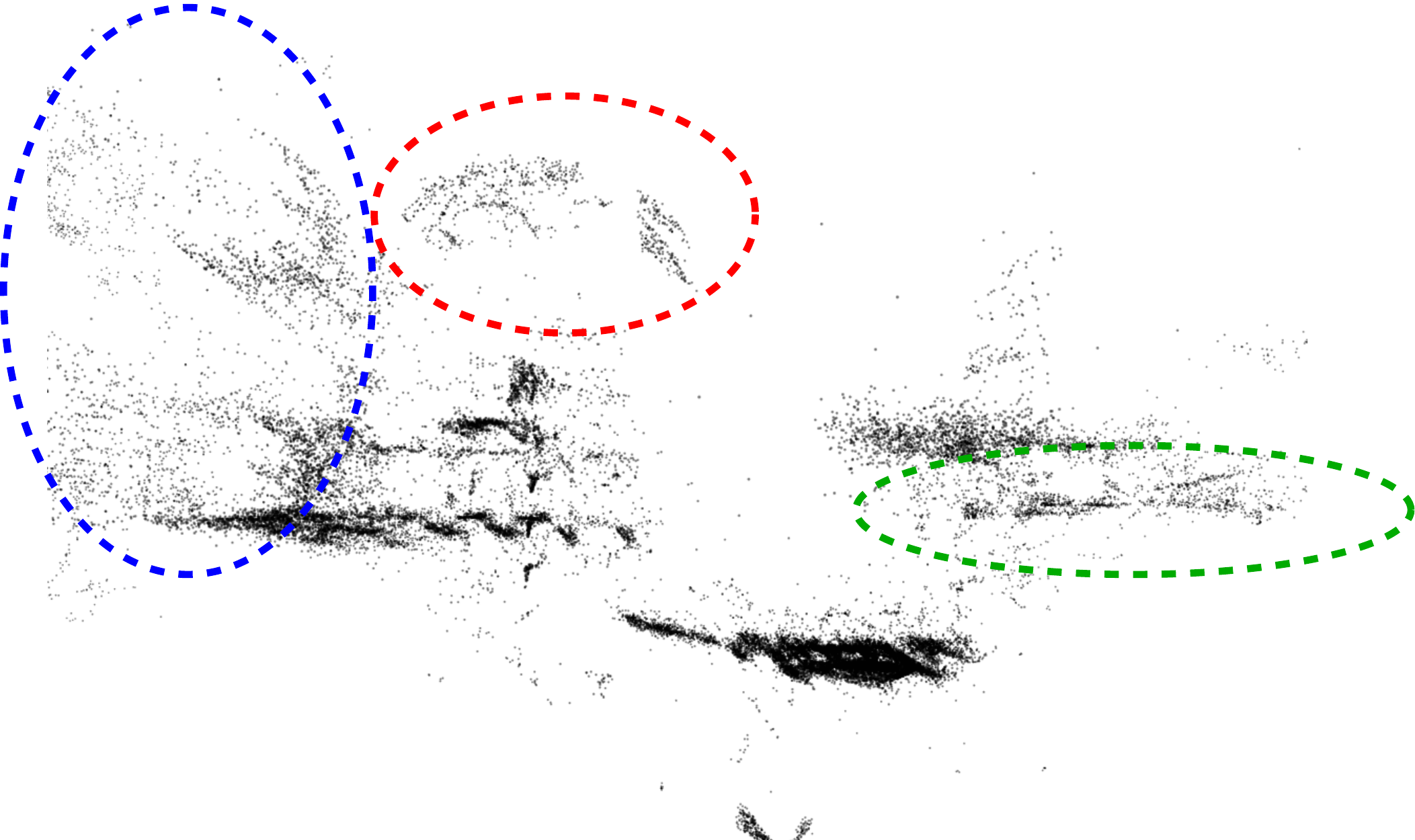}\\
        \caption{Top-down view -- \textbf{E} matrices only.}
    \end{subfigure}
    \caption{Visual comparison of the reconstructions of Yorkminster by \cite{sweeney2015theia} when initialized by the proposed (\textbf{E} + mult. \textbf{H}s (CC); top row) and traditional (\textbf{E} matrices; bottom) techniques. 
    {\color{blue}Blue} and {\color{ForestGreen}green} ellipses highlight areas that the proposed algorithm reconstructs significantly more accurately than the traditional approach. 
    The {\color{red}red} ellipse points to an erroneous area.
    ``CC'' stands for using the proposed sampler in the proposed method for multi-homography fitting. 
    }
    \label{fig:sfm-eval-Yorkminster}
\end{figure*}

\begin{figure*}
    \centering
    \begin{subfigure}[t]{0.492\textwidth}
   	 	\centering
        \includegraphics[width=1.0\columnwidth]{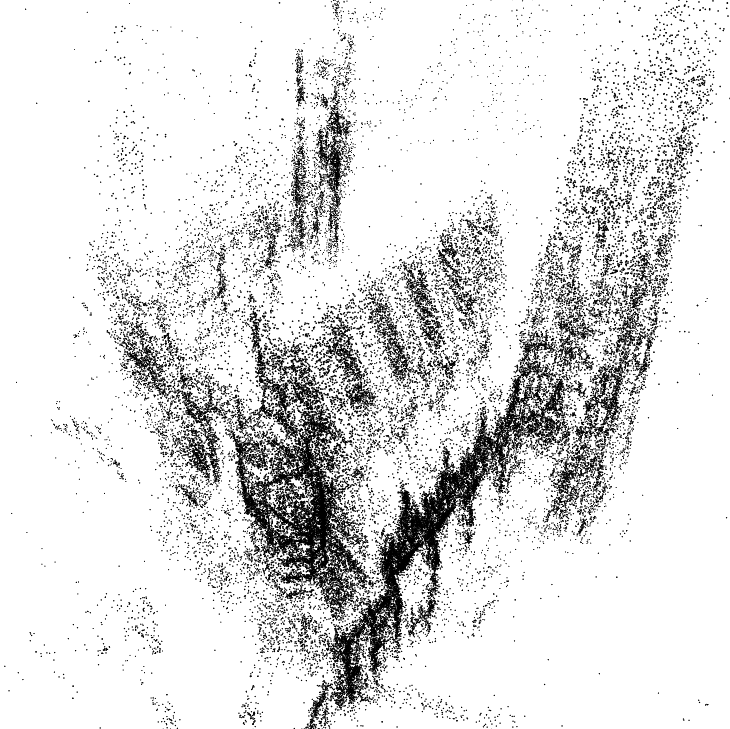}
        \caption{Frontal view -- \textbf{E} + mult. \textbf{H}s (CC).}
    \end{subfigure}%
    \begin{subfigure}[t]{0.492\textwidth}
   	 	\centering
        \includegraphics[width=1.0\columnwidth]{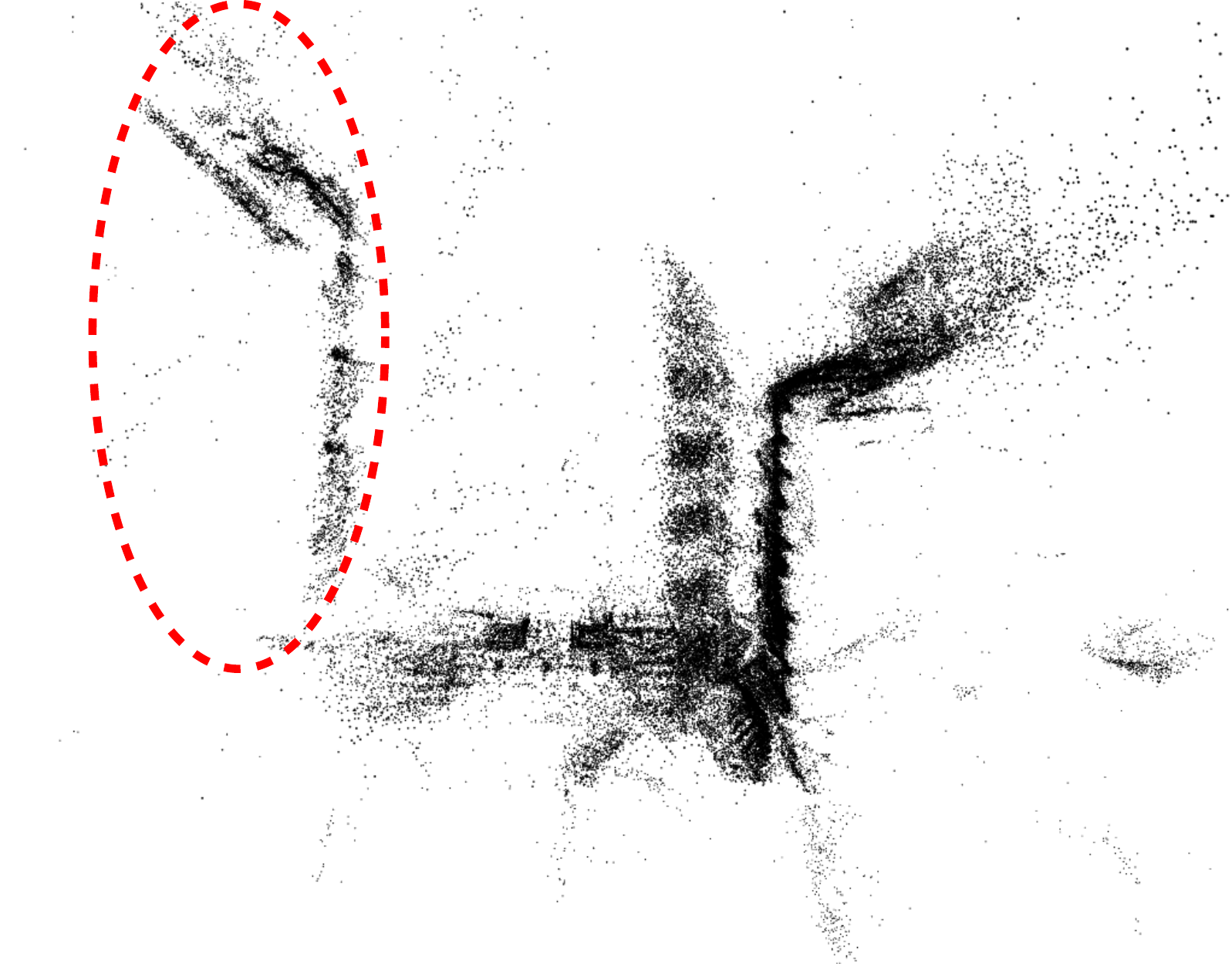}
        \caption{Top-down view -- \textbf{E} + mult. \textbf{H}s (CC).}
    \end{subfigure}
    \begin{subfigure}[t]{0.492\textwidth}
   	 	\centering
        \includegraphics[width=1.0\columnwidth]{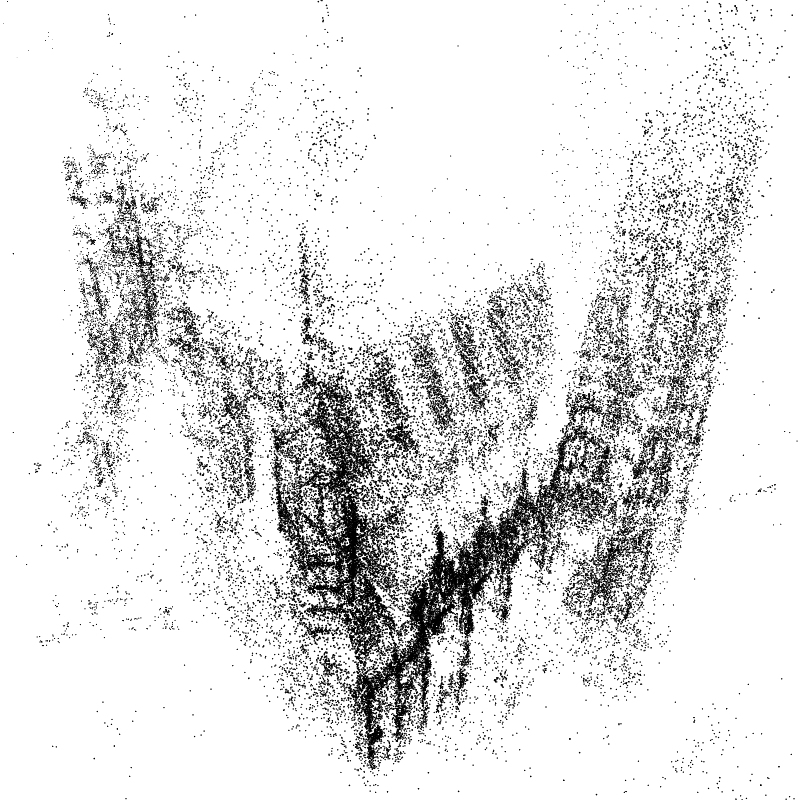}\\
        \caption{Frontal view -- \textbf{E} matrices only.}
    \end{subfigure}%
    \begin{subfigure}[t]{0.492\textwidth}
   	 	\centering
        \includegraphics[width=1.0\columnwidth]{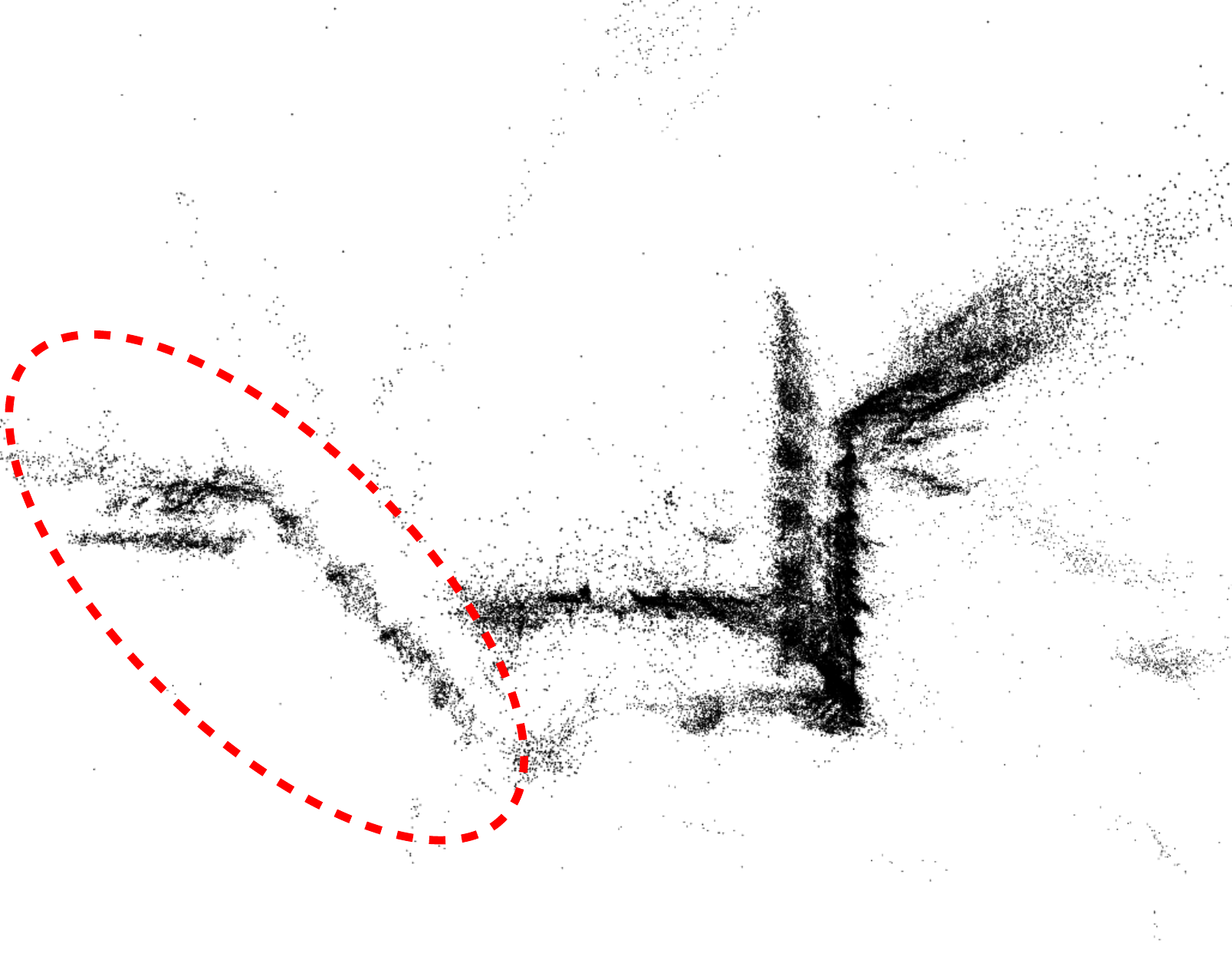}\\
        \caption{Top-down view -- \textbf{E} matrices only.}
    \end{subfigure}
    \caption{
    Visual comparison of the reconstructions of Vienna Cathedral by \cite{sweeney2015theia} when initialized by the proposed (\textbf{E} + mult. \textbf{H}s (CC); top) and traditional (\textbf{E} matrices; bottom) techniques. 
    The proposed approach preserves the parallelism of the walls of the cathedral ({\color{red}red} ellipse).
    ``CC'' stands for using the proposed sampler in the proposed method for multi-homography fitting.}
    \label{fig:sfm-eval-Vienna_Cathedral}
\end{figure*}

Additional visualizations are put in Figures~\ref{fig:sfm-eval-Yorkminster}~and~\ref{fig:sfm-eval-Vienna_Cathedral}, where the top rows show the results of \cite{sweeney2015theia} when initialized by a pose-graph estimated in the proposed way, exploiting an essential matrix and multiple homographies. 
The bottom rows show results when the pose-graph is estimated from essential matrices in the traditional way.
Colored ellipses mutually highlight parts of the two reconstructions with noticeable differences.
The traditional approach leads to reconstructions with fewer details and reduced precision compared to the proposed technique.

\begin{table*}[t]
\caption{Results of the global SfM algorithm from~\cite{sweeney2015theia} on the scenes from the 1DSfM dataset~\cite{wilson_eccv2014_1dsfm} when initialized by the pose-graph estimated from essential matrices ($\textbf{E}$ matrix), and the proposed method combined either with Progressive NAPSAC~\cite{barath2019pnapsac} or the proposed Connected Components (CC) samplers. As ground truth, we used reconstructions from COLMAP~\cite{schonberger2016structure}.
The averages and average medians of the rotation and position errors are reported in Table~\ref{tab:1DSfM}.}
\label{tab:1DSfM}
\centering
\resizebox{0.99\textwidth}{!}{
\setlength\aboverulesep{0pt}\setlength\belowrulesep{0pt}%
\setlength\extrarowheight{4.7pt}
\begin{tabular}{l r | S[detect-weight,table-format=3.0] S[detect-weight,table-format=6.0] |  S[detect-weight,table-format=2.2] S[detect-weight,table-format=2.2] S[detect-weight,table-format=2.2] | S[detect-weight,table-format=2.2] S[detect-weight,table-format=2.2] S[detect-weight,table-format=2.2] |S[detect-weight,table-format=1.2] S[detect-weight,table-format=1.2] S[detect-weight,table-format=1.2]}
\toprule
\rowcolor{black!10} 
 & &
   &
   &
  \multicolumn{3}{c|}{\cellcolor{black!10}orientation err (\SI{}{\degree})} &
  \multicolumn{3}{c|}{\cellcolor{black!10}position err (m)} &
  \multicolumn{3}{c}{\cellcolor{black!10}focal err\ ($\times 10^{-2}$)} \\
\hline
\rowcolor{black!10}
{} & {} & {$\#$ views} & {$\#$ tracks} & {AVG} & {MED} & {STD}  & {AVG} & {MED} & {STD} & {AVG} & {MED} & {STD}   \\ \midrule
\multirow{3}{*}{\rotatebox{90}{Alamo}\phantom{x}} & $\textbf{E}$ matrix & 493 & 104894 & \bfseries 2.46 & \bfseries 0.59 & 3.76 & \bfseries 1.60 & \bfseries 1.36 & \bfseries 3.98 & \bfseries 0.02 & \bfseries 0.01 & \bfseries 0.05 \\
& \textbf{E} + mult. \textbf{H}s & \bfseries 495 & \bfseries 110243 & 2.80 & 0.81 & 3.91 & 1.79 & 1.88 & 4.73 & \bfseries 0.02 & \bfseries 0.01 & \bfseries .05 \\
& \textbf{E} + mult. \textbf{H}s (CC) & 494 & 105920 & 2.59 & 0.62 & \bfseries 3.63 & 1.68 & 1.58 & 4.19 & \bfseries 0.02 & \bfseries 0.01 & \bfseries 0.05 \\ \midrule
\multirow{3}{*}{\rotatebox{90}{ Ellis Isl.\ }} & $\textbf{E}$ matrix & 211 & \bfseries 31200 & 4.21 & 2.90 & 4.69 & 5.59 & 3.43 & 10.57 & \bfseries 0.02 & \bfseries 0.01 & \bfseries 0.02 \\
& \textbf{E} + mult. \textbf{H}s & 210 & 30610 & \bfseries 3.49 & \bfseries 2.33 & \bfseries 3.00 & \bfseries 4.27 & \bfseries 3.09 & \bfseries 8.22 & \bfseries 0.02 & \bfseries 0.01 & \bfseries 0.02 \\
& \textbf{E} + mult. \textbf{H}s (CC) & \bfseries 215 & 31182 & 4.61 & 2.61 & 3.87 & 5.86 & 3.89 & 11.59 & \bfseries 0.02 & \bfseries 0.01 & \bfseries 0.02 \\ \midrule
\multirow{3}{*}{\rotatebox{90}{ Madrid M.\ }} & $\textbf{E}$ matrix & 299 & 56102 & 11.38 & 0.69 & 14.50 & 1.09 & 8.06 & 1.36 & \bfseries 0.06 & \bfseries 0.03 & \bfseries 0.10 \\
& \textbf{E} + mult. \textbf{H}s & \bfseries 327 & 50438 & \bfseries 4.00 & \bfseries 0.30 & \bfseries 5.63 & \bfseries 0.60 & \bfseries 2.86 & \bfseries 0.90 & 0.07 & \bfseries 0.03 & 0.14 \\
& \textbf{E} + mult. \textbf{H}s (CC) & 298 & \bfseries 57457 & 8.06 & 0.58 & 12.11 & 1.00 & 4.77 & 1.18 & \bfseries 0.06 & \bfseries 0.03 & \bfseries 0.10 \\ \midrule
\multirow{3}{*}{\rotatebox{90}{ Montreal }} & $\textbf{E}$ matrix & 432 & 106101 & \bfseries 1.34 & \bfseries 0.41 & 8.64 & \bfseries 0.82 & \bfseries 0.38 & \bfseries 1.22 & \bfseries 0.02 & \bfseries 0.01 & \bfseries 0.03 \\
& \textbf{E} + mult. \textbf{H}s & 435 & \bfseries 106498 & 1.52 & 0.46 & \bfseries 7.84 & 0.89 & 0.47 & 1.31 & \bfseries 0.02 & \bfseries 0.01 & \bfseries 0.03 \\
& \textbf{E} + mult. \textbf{H}s (CC) & \bfseries 436 & 104802 & 1.45 & 0.46 & 8.03 & 0.97 & 0.45 & 1.61 & \bfseries 0.02 & \bfseries 0.01 & \bfseries 0.03 \\ \midrule
\multirow{3}{*}{\rotatebox{90}{ NYC Lib.\ }} & $\textbf{E}$ matrix & 270 & \bfseries 57235 & 53.59 & 14.08 & \bfseries 3.86 & 14.10 & 52.95 & 7.26 & \bfseries 0.03 & \bfseries 0.01 & \bfseries 0.04 \\
& \textbf{E} + mult. \textbf{H}s & \bfseries 271 & 56435 & \bfseries 5.20 & \bfseries 2.94 & 3.96 & \bfseries 4.97 & \bfseries 4.23 & 6.73 & \bfseries 0.03 & \bfseries 0.01 & \bfseries 0.04 \\
& \textbf{E} + mult. \textbf{H}s (CC) & 270 & 55418 & 6.44 & 3.11 & 4.26 & 5.04 & 5.54 & \bfseries 6.59 & \bfseries 0.03 & \bfseries 0.01 & \bfseries 0.04 \\ \midrule
\multirow{3}{*}{\rotatebox{90}{ Piazza d.\ P.\ }} & $\textbf{E}$ matrix & \bfseries 291 & 42823 & 7.24 & 3.82 & 3.33 & 4.91 & 7.61 & 4.34 & \bfseries 0.03 & \bfseries 0.02 & \bfseries 0.04 \\
& \textbf{E} + mult. \textbf{H}s & 288 & \bfseries 44457 & 6.99 & 3.32 & 3.26 & 4.16 & 7.51 & 4.19 & \bfseries 0.03 & \bfseries 0.02 & 0.05 \\
& \textbf{E} + mult. \textbf{H}s (CC) & \bfseries 291 & 43510 & \bfseries 5.37 & \bfseries 2.53 & \bfseries 1.46 & \bfseries 3.28 & \bfseries 5.46 & \bfseries 3.54 & \bfseries 0.03 & \bfseries 0.02 & \bfseries 0.04 \\ \midrule
\multirow{3}{*}{\rotatebox{90}{ Piccadilly }} & $\textbf{E}$ matrix & \bfseries 1869 & 210821 & \bfseries \bfseries 4.71 & 0.35 & \bfseries 13.53 & 0.70 & 2.00 & \bfseries 1.05 & \bfseries 0.05 & \bfseries 0.03 & 0.15 \\
& \textbf{E} + mult. \textbf{H}s & 1656 & 141661 & 10.15 & 0.48 & 24.75 & 0.87 & 2.55 & 1.11 & \bfseries 0.05 & \bfseries 0.03 & \bfseries 0.14 \\
& \textbf{E} + mult. \textbf{H}s (CC) & 1860 & \bfseries 220045 & 4.96 & \bfseries 0.31 & 14.83 & \bfseries 0.66 & \bfseries 1.68 & \bfseries 1.05 & \bfseries 0.05 & \bfseries 0.03 & 0.15 \\ \midrule
\multirow{3}{*}{\rotatebox{90}{ Roman F.\ }} & $\textbf{E}$ matrix & 989 & \bfseries 208457 & 4.87 & \bfseries 14.76 & 4.68 & \bfseries 22.25 & 3.86 & 82.77 & \bfseries 0.03 & \bfseries 0.02 & \bfseries 0.07 \\
& \textbf{E} + mult. \textbf{H}s & 991 & 204432 & \bfseries 4.56 & 15.64 & \bfseries 3.37 & 22.90 & \bfseries 3.78 & 82.49 & \bfseries 0.03 & \bfseries 0.02 & \bfseries 0.07 \\
& \textbf{E} + mult. \textbf{H}s (CC) & \bfseries 995 & 206641 & 4.85 & 15.78 & 3.59 & 23.61 & 4.01 & \bfseries 82.29 & \bfseries 0.03 & \bfseries 0.02 & \bfseries 0.07 \\ \midrule
\multirow{3}{*}{\rotatebox{90}{ Tower }} & $\textbf{E}$ matrix & \bfseries 406 & \bfseries 96481 & 6.03 & \bfseries 9.48 & 12.55 & \bfseries 25.04 & \bfseries 2.42 & \bfseries 38.79 & \bfseries 0.02 & \bfseries 0.01 & \bfseries 0.03 \\
& \textbf{E} + mult. \textbf{H}s & 397 & 95394 & \bfseries 5.29 & 10.58 &\bfseries 6.29 & 26.47 & 3.39 & 40.70 & \bfseries 0.02 & \bfseries 0.01 & \bfseries 0.03 \\
& \textbf{E} + mult. \textbf{H}s (CC) & 405 & 96088 & 5.83 & 10.94 & 8.87 & 26.56 & 3.54 & 40.54 & \bfseries 0.02 & \bfseries 0.01 & \bfseries 0.03 \\ \midrule
\multirow{3}{*}{\rotatebox{90}{ Trafalgar }} & $\textbf{E}$ matrix & \bfseries 4111 & \bfseries 354494 & 18.03 & 16.79 & 32.09 & 23.92 & 10.70 & \bfseries 29.63 & \bfseries 0.02 & \bfseries 0.01 & \bfseries 0.03 \\
& \textbf{E} + mult. \textbf{H}s & 4097 & 349621 & 19.10 & \bfseries 16.14 & 41.86 & \bfseries 23.74 & \bfseries 7.20 & 30.73 & \bfseries 0.02 & \bfseries 0.01 & \bfseries 0.03 \\
& \textbf{E} + mult. \textbf{H}s (CC) & 4088 & 349784 & \bfseries 18.00 & 17.09 & \bfseries 31.97 & 24.79 & 10.93 & 30.64 & \bfseries 0.02 & \bfseries 0.01 & \bfseries 0.03 \\ \midrule
\multirow{3}{*}{\rotatebox{90}{ Vienna C.\ }} & $\textbf{E}$ matrix & 705 & 160363 & 14.47 & 7.49 & 9.86 & 10.96 & 9.40 & 11.55 & \bfseries 0.02 & \bfseries 0.01 & \bfseries 0.05 \\
& \textbf{E} + mult. \textbf{H}s & 612 & 92051 & 26.35 & 13.79 & 29.71 & 22.85 & 13.10 & 25.45 & \bfseries 0.02 & \bfseries 0.01 & \bfseries 0.05 \\
& \textbf{E} + mult. \textbf{H}s (CC) & \bfseries 707 & \bfseries 160503 & \bfseries 4.72 & \bfseries 6.97 & \bfseries 4.84 & \bfseries 10.15 & \bfseries 3.18 & \bfseries 11.03 & \bfseries 0.02 & \bfseries 0.01 & \bfseries 0.05 \\ \midrule
\multirow{3}{*}{\rotatebox{90}{ Yorkmins.\ }} & $\textbf{E}$ matrix & 399 & 98396 & 5.52 & 7.61 & 3.57 & 12.13 & 4.99 & 17.70 & \bfseries 0.03 & \bfseries 0.01 & \bfseries 0.04 \\
& \textbf{E} + mult. \textbf{H}s & \bfseries 402 & 100985 & 5.68 & 7.74 & 3.46 & 12.68 & 5.11 & 20.03 & \bfseries 0.03 & \bfseries 0.01 & \bfseries 0.04 \\
& \textbf{E} + mult. \textbf{H}s (CC) & 399 & \bfseries 109132 & \bfseries 3.49 & \bfseries 6.27 & \bfseries 2.90 & \bfseries 11.26 & \bfseries 2.91 & \bfseries 17.12 & \bfseries 0.03 & \bfseries 0.01 & \bfseries 0.04 \\
\bottomrule
\end{tabular}}
\end{table*}

\section{Translation from Known Rotation}

In Section 4.2., we propose to estimate the relative pose from multiple homographies and the essential matrix by decomposing them and choosing the pose that leads to the most inliers when thresholding the re-projection error. 
We found that, while the estimated rotation matrix often is accurate, the translation can be improved by re-estimating it from the found inliers considering the known rotation. 



In this section, we briefly describe the translation estimation procedure given a known rotation matrix.
It is well-known~\cite{hartley2003multiple} that the essential matrix is defined as 
\begin{equation*}
    \matr E =\left[\matr t\right]_{\times}\matr R,
\end{equation*}
where $\matr t \in \mathbb{R}^3$ and $\matr R \in \text{SO}(3)$ are, respectively, the translation vector and rotation matrix, and $\left[\matr t\right]_{\times}$ is the cross-product matrix of $\matr t$ as follows:
\begin{equation*}
    \left[\matr t\right]_{\times} = \left[\begin{array}{ccc}
0 & -t_{z} & t_{y}\\
t_{z} & 0 & -t_{x}\\
-t_{y} & t_{x} & 0
\end{array}\right].
\end{equation*}
Essential matrix $\matr E$ describes the relationship of a point correspondence in the images via the well-known epipolar constraint as follows: 
\begin{equation*}
    \matr p_2^\text{T} \matr E \matr p_1 = 0, 
\end{equation*}
where $\matr p_1 = \left[ u_1 \; v_1 \; w_1 \right]^\text{T}$ and $\matr p_2 =  \left[ u_2 \; v_2 \; w_2 \right]^\text{T}$ are homogeneous points in the normalized image plane, \ie, normalized by the intrinsic camera matrices.
Considering $\matr R$ to be known, we are given the following constraint
\begin{equation*}
    \matr p_2^\text{T} \left[\matr t\right]_{\times}\matr R \matr p_1 = 0, 
\end{equation*}
where the only unknowns are the three translation components $\matr t = \left[ t_x \; t_y \; t_z \right]^\text{T}$. Multiplication $\matr R \matr p_1$ can be pre-calculated as $\matr p_1' = \matr R \matr p_1$. 
Formula $\matr p_2^\text{T} \left[\matr t\right]_{\times}\matr p_1'$ leads to:
\begin{equation}
\begin{aligned}
-u_2t_{z}v_1'+u_2t_{y}w_1'+v_2t_{z}u_1'- 
v_2t_{x}w_1'-w_2t_{y}u_1'+w_2t_{x}v_1' = 0.
\end{aligned}
\label{eq:epiconstraint}
\end{equation}
%
Eq.~\ref{eq:epiconstraint} is linear in the elements of the translation vector. %
Therefore, the equation can be reformulated as
\begin{equation*}
\left[\begin{array}{c}
{v_1'}w_2-{w_1'}v_2\\
u_2{w_1'}-w_2{u_1'}\\
v_2{u_1'}-u_2{v_1'}
\end{array}\right]^{\text{T}}\left[\begin{array}{c}
t_{x}\\
t_{y}\\
t_{z}
\end{array}\right]=0.
\end{equation*}
If at least two point correspondences are given, a homogeneous linear system of equations is obtained. The optimal solution, in the LSQ sense, is given via calculating the null-vector of the coefficient matrix. 

\section{Trajectories of Fast-moving Objects}

We show example visualizations of trajectory estimation of fast-moving objects in Figure~\ref{fig:fmosupp}. 
After extracting blur kernels that encode the object motion, we apply a multi model fitting algorithm recovering line segments. 
The estimated line segments are colored in red. 
The ground truth line segments are generated by applying a classical state-of-the-art object tracking algorithm on high-speed camera footage with manual annotations, which is shown in green. 
We show the results of sequential RANSAC as originally proposed in~\cite{tbd}. 
Additionally, we show final trajectories after filtering and refinement by~\cite{tbd}. 
Quantitative results are reported in the paper.

Notice that the line segments found by seq.\ RANSAC are not continuous, \ie, there is a clear gap between all of them. 
This is caused by the hard point-to-line assignment used in seq.\ RANSAC and in the state-of-the-art multi-model fitting algorithms. 
Using the proposed method allows finding continuous chains that lead to better trajectories as shown in the last column and, also, in Table 3 in the main paper.

\newcommand{\addimg}[1]{\includegraphics[width=0.165\linewidth]{suppl/lines/#1.png}}
\newcommand{\addex}[1]{\addimg{im0#1} & \addimg{psf0#1}}
\newcommand{\addimgnew}[2]{\includegraphics[trim=#2,clip,width=0.165\linewidth]{suppl/linesnew/0#1.png}}
\newcommand{\addexnew}[2]{\addimgnew{#1_imgt}{#2} & \addimgnew{#1_psf}{#2} & \addimgnew{#1_imseqran}{#2} & \addimgnew{#1_imseqran_final}{#2} & \addimgnew{#1_improgxp}{#2} & \addimgnew{#1_improgxp_ref}{#2}}
\begin{figure*}
\centering
\setlength{\tabcolsep}{0.05em} 
\renewcommand{\arraystretch}{0.25} 
\begin{tabular}{@{}cccccc@{}}
\addexnew{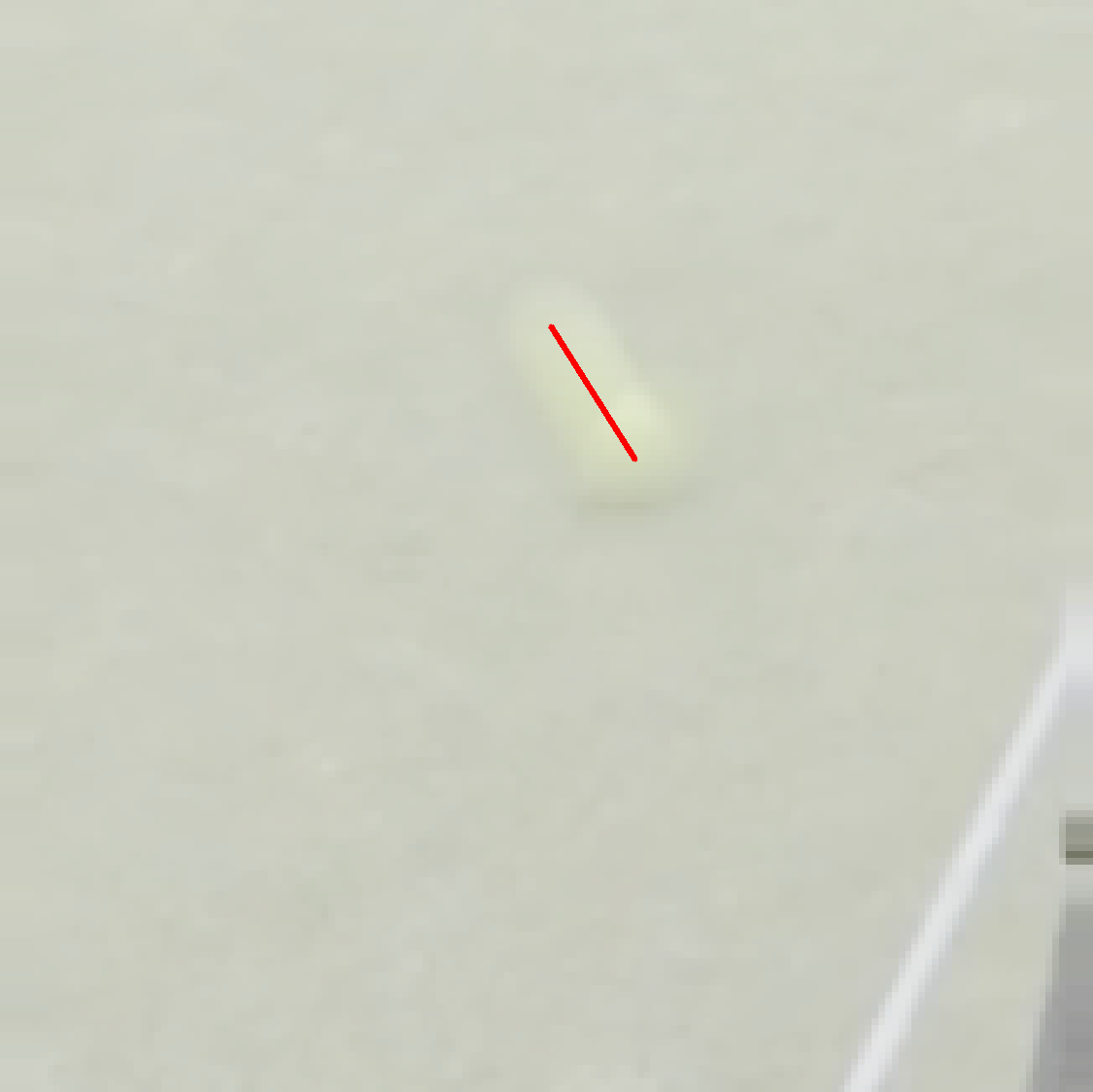}{180 220 100 100} \\
\addexnew{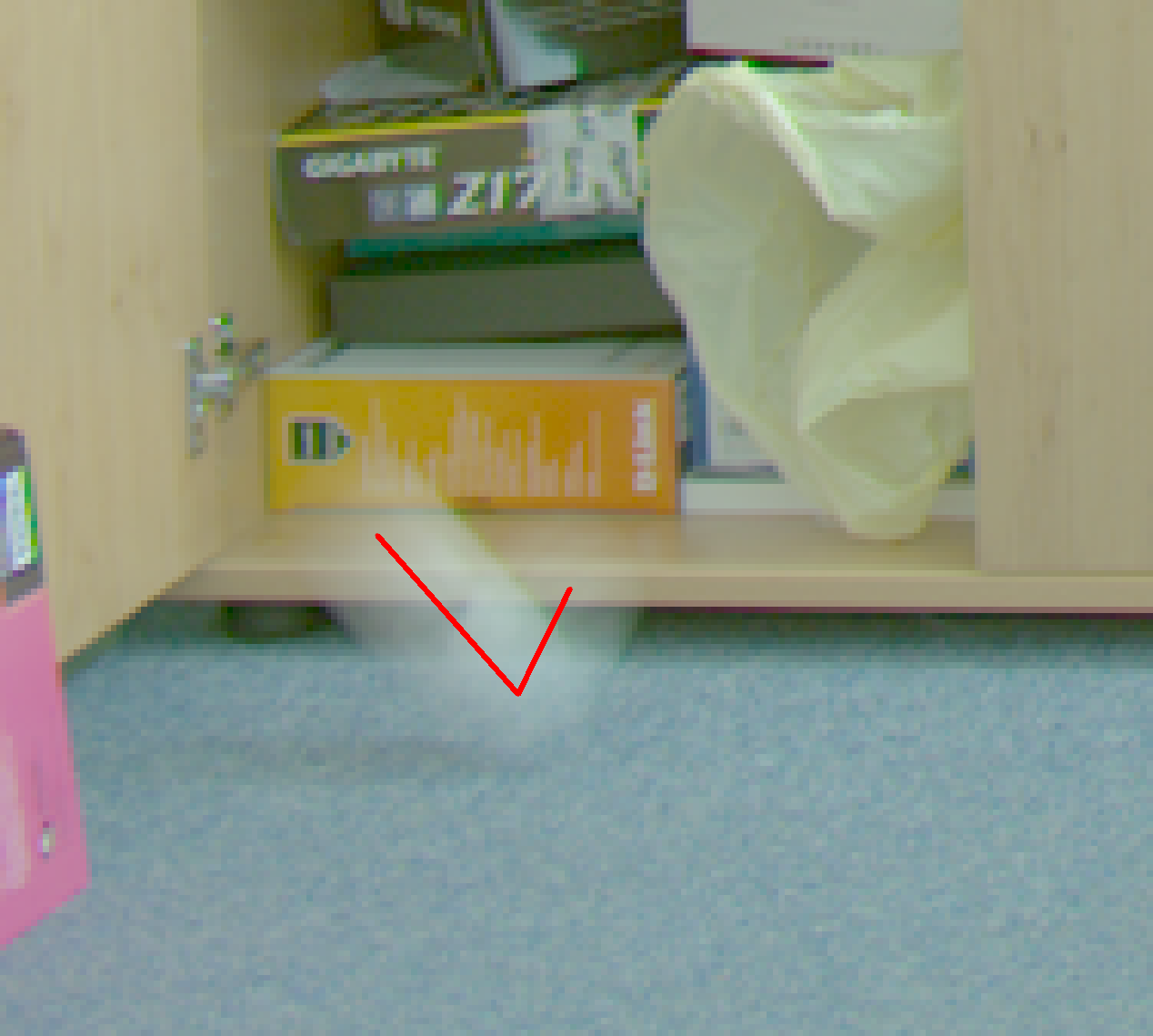}{100 100 150 150} \\
\addexnew{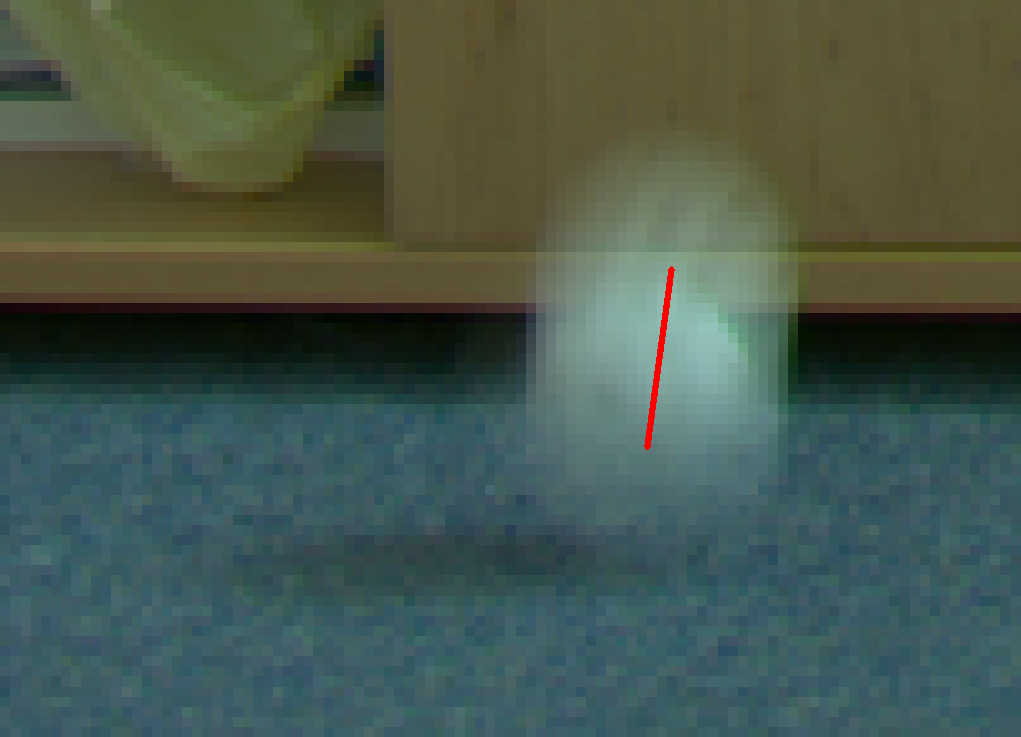}{100 0 0 0} \\
\addexnew{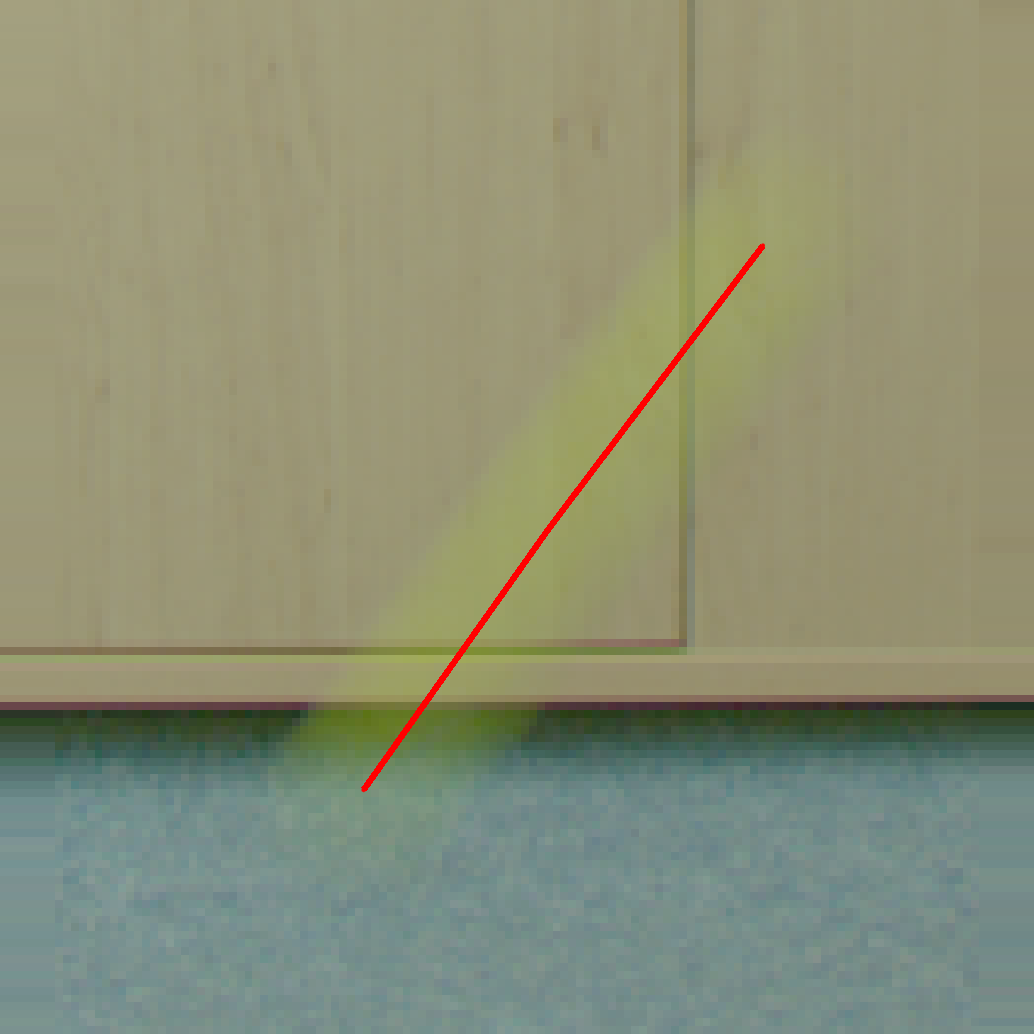}{0 0 0 0} \\
\addexnew{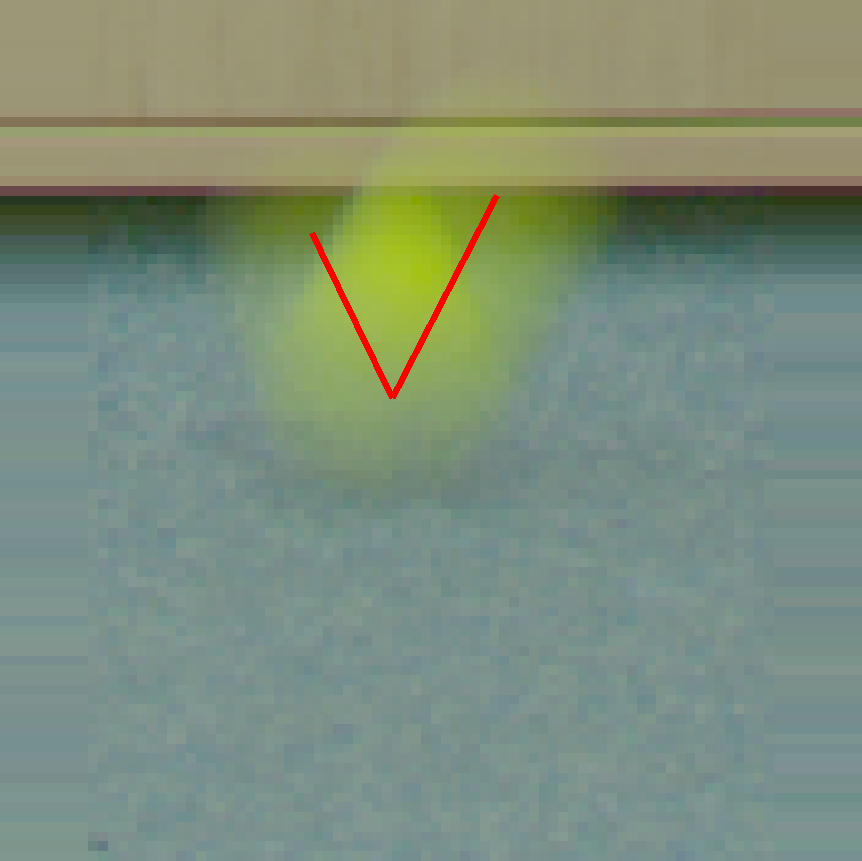}{0 100 0 0} \\
Input image & Blur kernel & Seq. RANSAC & + filter \& refine & Proposed & + filter \& refine \\
\end{tabular}
\caption{Fitting multiple line segments for trajectory estimation of fast-moving objects. The estimated and ground truth segments are colored by red and green, respectively. The original Tracking by Deblatting~\cite{tbd} method for trajectory estimation of fast-moving objects uses the sequential RANSAC algorithm. Therefore, we report results using their implementation. The filtering and refinement are done by the method proposed in~\cite{tbd}.
After post-processing by filtering and refinement, the results from the proposed algorithm more often cover the sought trajectory than by the other methods. The results of seq.\ RANSAC, besides being qualitatively worse, \ie\ missing a segment in rows 1, 3, and 5, suffer from the single-model  assignment of inliers which shows as a gap between consecutive segments. The width of the gap equals to the inlier threshold of seq.\ RANSAC.
}
\label{fig:fmosupp}
\end{figure*}

{\small
\bibliographystyle{ieee_fullname}
\bibliography{egbib}
}

\end{document}